\documentclass[journal]{IEEEtran}

\usepackage{amsthm}
\usepackage{amssymb}
\usepackage{amsmath}
\usepackage{amsfonts}

\usepackage{cite}
\usepackage{float}
\usepackage{graphicx}
\usepackage{mwe}
\usepackage{graphbox}
\usepackage{setspace}
\usepackage{epstopdf}
\usepackage{epsf}
\usepackage{enumitem}
\usepackage{hyperref}

\usepackage{bm}
\usepackage{acronym}

\usepackage[ruled,vlined]{algorithm2e}

\usepackage[dvipsnames]{xcolor}

\theoremstyle{remark}	\newtheorem{theorem}{Theorem}
\theoremstyle{remark}	
\theoremstyle{remark}	

\begin{document}

\title{Convolutional Sparse Coding Fast Approximation with Application to Seismic Reflectivity Estimation} 

\author{Deborah Pereg, Israel Cohen, and Anthony A. Vassiliou
\thanks{D. Pereg and I. Cohen are with Andrew and Erna Viterby Faculty of Electrical and Computer Engineering, Technion -- Israel Institute of Technology,
        Technion City, Haifa 3200003, Israel (e-mail: deborahp@campus.technion.ac.il; icohen@ee.technion.ac.il).}
\thanks{A.~A. Vassiliou is with GeoEnergy, Inc., United States (e-mail: anthony@geoenergycorp.com). }
}

\maketitle

\begin{abstract}
In sparse coding, we attempt to extract features of input vectors, assuming that the data is inherently structured as a sparse superposition of basic building blocks. Similarly, neural networks perform a given task by learning features of the training data set. Recently both data driven and model driven feature extracting methods have become extremely popular and have achieved remarkable results. Nevertheless, practical implementations are often too slow to be employed in real life scenarios, especially for real time applications. We propose a speed-up upgraded version of the classic iterative thresholding algorithm, that produces a good approximation of the convolutional sparse code within 2-5 iterations. The speed advantage is gained mostly from the observation that most solvers are slowed down by inefficient global thresholding. The main idea is to normalize each data point by the local receptive field energy, before applying a threshold. This way, the natural inclination towards strong feature expressions is suppressed, so that one can rely on a global threshold that can be easily approximated, or learned during training. The proposed algorithm can be employed with a known predetermined dictionary, or with a trained dictionary. The trained version is implemented as a neural net designed as the unfolding of the proposed solver. The performance of the proposed solution is demonstrated via the seismic inversion problem in both synthetic and real data scenarios. We also provide theoretical guarantees for a stable support recovery. Namely, we prove that under certain conditions the true support is perfectly recovered within the first iteration.
\end{abstract}
\maketitle


\section{Introduction}

In sparse coding, we attempt to decompose an observation signal $\mathbf{y} \in \mathbb{R}^{N \times 1}$ into its building blocks (atoms) \cite{Chen}. Namely, the sparse representations model \cite{Elad} assumes a signal $\mathbf{y}\in\mathbb{R}^{N \times 1}$ that can be formulated as a sparse superposition of atoms. Mathematically speaking, we assume the observation signal $\mathbf{y}$ obeys
\begin{equation}\label{1.1}
\mathbf{y}=\mathbf{D}\mathbf{x},
\end{equation}
where $\mathbf{D}\in\mathbb{R}^{N \times M}$ is a matrix called the dictionary, that consists of the atoms $\mathbf{d}_i \in\mathbb{R}^{N \times 1}, \ i=1,...,M$, as its columns, and
$\mathbf{x} \in \mathbb{R}^{M \times 1}$ is the \textit{sparse} vector of the atoms weights.
Generally, we do not impose any relation between $N$ and $M$. That is, the dictionary could either be complete ($N=M$), over-complete ($N<M$) or under-complete ($N>M$).

Over the years, great efforts have been invested in finding sparse solutions to (\ref{1.1}). Breaking down a signal into its building blocks has become a popular task in many fields related to signal processing, such as: image processing \cite{Elad}, computer vision \cite{Jarrett:2009},  compressed sensing \cite{Donoho}, radar \cite{Baranuik}, ultrasound imaging \cite{Bendory2}, seismology \cite{Taylor,Nguyen,Zhang2}, visual neurosciense \cite{Olshausen:1996,Lee:2009}, and more. Moreover, from a certain perspective, neural networks can be viewed as an unfolding of an iterative sparse coding solver \cite{Papyan:2017}.  In a sense, deep neural nets (DNNs) are trained to seek the atoms weights, specifically convolutional neural networks (CNNs) \cite{Lecun:1998}, that are trained to find a set of filters tailored to perform a classification or a regression task at hand. However, in many real-time applications, such as pattern recognition, sparse coding is still a bottleneck in terms of inference time. Most often, a sparse code is required for every image patch.
Consequently, many attempts have been made to pursue faster methods for sparse coding.

The Iterative shrinkage thresholding algorithm (ISTA) \cite{Daubechies:2004} is one of the most popular algorithms for sparse coding. Despite its simplicity, ISTA is considered as a slow algorithm. Over time, faster extensions have been suggested, such as, Fast-ISTA (FISTA) \cite{Beck:2009}, Learned-ISTA (LISTA)\cite{Lecun:2010} and Ada-LISTA \cite{Aberdam:2020}. LISTA \cite{Lecun:2010}, for example, uses a learned substitute dictionary, and ada-LISTA \cite{Aberdam:2020} incorporates an adaptive threshold, where the first threshold corresponds to the maximal feature weight in the data. Then, the thresholds are gradually decreasing at each iteration. 
Since the seminal idea of LISTA - to unroll the iterative algorithm into feed-forward layers - was first proposed,  many similar sparse coding model-based deep learning methods have been proposed, such as: ADMM with CNN \cite{Zhao:2018}, ADMM-CSNet \cite{Yang:2020}, and FISTA-Net \cite{Xiang:2021}. These methods are designed to provide accurate and fast reconstruction compared with other deep learning methods.

Inspired by the classic iterative thresholding algorithms, in this work, we propose a fast alternative algorithm that produces a good approximation of a convolutional sparse code. Most solvers are slowed down by the use of one global threshold (bias) to detect each local feature shift along the signal, or a predetermined constant local threshold. This way, even if the signal (or the input batch) is normalized, when we apply a threshold at each iteration, if the threshold is too high, weak expressions are annihilated, and strong expressions can ``cast a shadow'' over low-energy regions in the signal, which can be interpreted as false-positive support locations. On the other hand, if the threshold is very small, as often is the case in ISTA, many iterations are required to compensate for false detections in early iterations, especially in the presence of noise, and in real-time applications due to model perturbations.

Alternatively, we propose to normalize each data point by a locally focused data energy measure, before applying a threshold. In other words, each receptive field of the data is scaled with respect to the local energy. This way even when the data is inherently unbalanced, we can still use a common bias for all receptive fields, without requiring many iterations that are usually required in order to globally detect the features support. The proposed algorithm can be employed with a known predetermined dictionary and global fixed bias terms for each iteration, or with a learned dictionary and learned bias terms. In practice, the trained version is implemented as a small recurrent CNN that corresponds to a few unfolded iterations of the proposed method. An approximate solution is produced within only 2-5 iterations, that is, to our knowledge, the state of the art performance in terms of speed and computational complexity versus accuracy error. 

We further demonstrate the applicability of the proposed solution to seismic inversion, via experimental results with real data and synthetic data, demonstrating that even if the mutual coherence of the dictionary is relatively high, the first iteration of the method accurately detects at least 70$\%$ of the features weights. The following few iterations make the required corrections. We performed extensive synthetic data and real data numerical experiments, in order to verify the robustness of the method in noisy and attenuating environment. We also prove that under sufficient separation and sufficiently low mutual coherence the first iteration of our method is guaranteed to perfectly recover the true support. Furthermore, in our opinion, the proposed predictor can be potentially included in learning systems in many applications, such as, recognition systems and biomedical-imaging super-resolution.

The main contribution of this work is a highly efficient method for fast convolutional sparse coding approximation. We also propose a learning-based (data driven) variation of the proposed method. The  performance is demonstrated via extensive numerical experiments conducted with seismic real data as well as with synthetic data.
Furthermore, we prove that the proposed method achieves a reliable solution under sufficient separation or low mutual coherence.

The remainder of this paper is organized as follows. Section~\ref{sec2} provides the necessary background for sparse representations, convolutional sparse coding and iterative thresholding algorithms. Section~\ref{sec3} describes the seismic inversion problem. 
In Section~\ref{sec4}, we present the proposed method and its theoretical guarantees.
Sections~\ref{sec5}-\ref{sec6} introduce synthetic and real data experimental results.
Section~\ref{sec7} proposes a learned version of the proposed algorithm, and demonstrates its employment experimentally. Finally, Section~\ref{sec8} concludes and discusses future research directions.

\section{Background and Related Work}
\label{sec2}
\subsection{Sparse Representations}
The sparse representations model \cite{Elad} assumes a signal $\mathbf{y}\in\mathbb{R}^{N \times 1}$ that is formulated as a sparse superposition of atoms:
\begin{equation}
\mathbf{y}=\mathbf{D}\mathbf{x}, \tag{\ref{1.1}}
\end{equation}
where $\mathbf{D}\in\mathbb{R}^{N \times M}$ is a matrix called the dictionary, built of the atoms $\mathbf{d}_i \in\mathbb{R}^{N \times 1}, \ i=1,...,M$, as its columns, and
$\mathbf{x} \in \mathbb{R}^{M \times 1}$ is the sparse vector of the atoms weights. 

An immense amount of work has been dedicated to sparse coding, that is, to the recovery of $\mathbf{x}$. To find the sparsest solution, the one with the smallest $\ell_0$-norm, we attempt to solve
\begin{equation}\label{1.2}
(P_0): \qquad \min_\mathbf{x} \|\mathbf{x}\|_0 \qquad 	\mathrm{s.t.}	 \qquad 	\mathbf{y}=\mathbf{D}\mathbf{x},
\end{equation}
where $\|\mathbf{x}\|_0$ denotes the number of non-zeros in $\mathbf{x}$.
Since $P_0$ has been proven to be, in general, NP-Hard \cite{Natarajan:1995}, we often replace the $\ell_0$-norm with the $\ell_1$-norm
 \begin{equation}\label{1.3a}
(P_1): \qquad \min_\mathbf{x} \|\mathbf{x}\|_1 \qquad 	\mathrm{s.t.}	 \qquad 	\mathbf{y}=\mathbf{D}\mathbf{x},
\end{equation}
where $\|\mathbf{x}\|_1 \triangleq \sum_i |x_i|$.
In noisy environment or when some error is allowed we attempt to solve 
\begin{equation}\label{1.3b}
(P_{1,\varepsilon}): \qquad \min_\mathbf{x} \|\mathbf{x}\|_1 \qquad 	\mathrm{s.t.}	 \qquad 	 \|\mathbf{y}-\mathbf{D}\mathbf{x}\|_2 \leq \varepsilon,
\end{equation}
where $\|\mathbf{x}\|_2 \triangleq \sqrt{\sum_i x^2_i}$.
Under certain conditions, the sparsest solution to $P_0$ and $P_1$ has been proven to be unique and can be retrieved using practical algorithms, such as orthonormal matching pursuit (OMP) or basis pursuit (BP), depending on the dictionary's properties and the sparsity of $\mathbf{x}$. Namely, under the assumption that $\| \mathbf{x} \|_0 < \frac{1}{2}\Big( 1+ \frac{1}{\mu(\mathbf{D})} \Big)$,
where $\mu(\mathbf{D})$ is the mutual coherence defined as the maximal correlation coefficient between two dictionary atoms,
\begin{equation}\label{1.4}
\mu(\mathbf{D})=\max_{i \neq j} \frac{\Big|\mathbf{d}^T_i\mathbf{d}_j\Big|}{\|\mathbf{d}_i\|_2 \cdot \|\mathbf{d}_j\|_2},
\end{equation}
the true sparse code $\mathbf{x}$ can be perfectly recovered \cite{candes:2011}.

One of the most intuitive ways to recover $\mathbf{x}$ is to project $\mathbf{y}$ on the dictionary, and then extract the atoms with the strongest response by taking a hard or a soft threshold. In other words, the solution is a closed-form solution, formulated as 
$\mathbf{x}=\mathcal{H}_\beta(\mathbf{D}^T\mathbf{y})$ or $\mathbf{x}=\mathcal{S}_\beta(\mathbf{D}^T\mathbf{y})$,
where the hard threshold and the soft threshold operators are respectively defined as
\begin{equation*}
\mathcal{H}_\beta(z)=  
\begin{cases}
z, & |z| > \beta \\
0, & |z| \leq \beta\\
\end{cases},
\end{equation*}
and 
\begin{equation*}
\mathcal{S}_\beta(z)=  
\begin{cases}
z+\beta, & z<-\beta \\
0, &  |z| \leq \beta\\
z-\beta, & z>\beta
\end{cases}.
\end{equation*}
Note that the rectified linear units (ReLU) activation function commonly used in DNNs satisfies
\begin{equation*}
\mathrm{ReLU}(z-\beta)= \max (z-\beta,0) =\mathcal{S}^{+}_\beta(z) \triangleq 
\begin{cases}
0, &  z \leq \beta\\
z-\beta, & z>\beta
\end{cases}.
\end{equation*}
Therefore, the soft threshold solution can be also written as
\begin{flalign*}\label{1.5}
\mathbf{x}
& =\mathcal{S}^{+}_\beta(\mathbf{D}^T\mathbf{y})-\mathcal{S}^{+}_\beta(-\mathbf{D}^T\mathbf{y})
\\ 
& 
= \mathrm{ReLU}(\mathbf{D}^T\mathbf{y}-\beta)-\mathrm{ReLU}(-\mathbf{D}^T\mathbf{y}-\beta).
\end{flalign*}
However, simple thresholding is guaranteed to recover the true support only under a more restrictive assumption that $\| \mathbf{x} \|_0 < \frac{1}{2} \Big( 1+ \frac{1}{\mu(\mathbf{D})} \frac{|\mathbf{x}|_{\mathrm{min}}}{|\mathbf{x}|_{\mathrm{max}}} \Big)$, where $|\mathbf{x}|_{\mathrm{min}}$, and $|\mathbf{x}|_{\mathrm{max}}$ are the minimum and maximum values of the vector $|\mathbf{x}|$ on the support, implying that when the data is unbalanced - this approach is bound to collapse.

Papyan et al. \cite{Papyan:2017} show that the forward pass of CNNs is equivalent to the layered thresholding algorithm designed to solve the convolutional sparse coding (CSC) problem.
A convolutional layer's forward pass is guaranteed to recover a sparse estimate of the underlying representations of an input signal, only for dictionaries with very low mutual coherence, and when the ratio between the maximal and minimal spikes' ampiltudes in absolute value is close to one. In addition, the $\ell_{0,\infty}$ of the true solution $\mathbf{x}$, that is, the required maximal number of non-zeros in a stripe of coefficients contributing to a data point, depends on the ratio $\frac{|\mathbf{x}|_{\mathrm{min}}}{|\mathbf{x}|_{\mathrm{max}}}$. Therefore, under these very limiting conditions, when applied to real-life applications, such as the seismic reflectivity estimation problem, for which the mutual coherence of the dictionary is very high and the seismic data set is inherently unbalanced (i.e., with peak and trough amplitudes that are not close), this course of action is inadequate. 

One may also wonder whether the dictionary $\mathbf{D}$ is known, and if not, then how and under what conditions could one find the atoms or features concerning his or her specific problem. Some of these questions are addressed in Section~\ref{sec7} and in Appendices A and C.

\subsection{Iterative Shrinkage Algorithms}
Consider the cost function
\begin{equation*}
f(\mathbf{x})= \frac{1}{2}\left\| \mathbf{y} - \mathbf{D} \mathbf{x} \right\|_2^2 + \lambda {\left\| {\mathbf{x}}\right\|_1}.
\end{equation*}
Following Majorization Minimization (MM) strategy, we can build a surrogate function \cite{Daubechies:2004,Elad}
\begin{flalign*}
& Q(\mathbf{x},\mathbf{x}_\theta) = f(\mathbf{x}) + d(\mathbf{x},\mathbf{x}_\theta) = \\
& \frac{1}{2}\left\| \mathbf{y} - \mathbf{D} \mathbf{x} \right\|_2^2 
+
 \lambda {\left\| {\mathbf{x}}\right\|_1} +
\frac{c}{2} \|\mathbf{x}-\mathbf{x}_\theta \|_2^2-\frac{1}{2} \| \mathbf{D} \mathbf{x} -   \mathbf{D} \mathbf{x}_\theta\|_2^2.
\end{flalign*}
The parameter $c$ is chosen such that the added expression 
\begin{equation*}
d(\mathbf{x},\mathbf{x}_\theta)=Q(\mathbf{x},\mathbf{x}_\theta)-f(\mathbf{x})= \frac{c}{2} \|\mathbf{x}-\mathbf{x}_\theta \|_2^2-\frac{1}{2} \| \mathbf{D} \mathbf{x} -   \mathbf{D} \mathbf{x}_\theta\|_2^2
\end{equation*}
is strictly convex, requiring its Hessian to be positive definite, $c \mathbf{I} - \mathbf{D}^T\mathbf{D} \prec \mathbf{0}$. Therefore $ c > \| \mathbf{D}^T\mathbf{D} \|_2 = \lambda_{max}(\mathbf{D}^T\mathbf{D})$, i.e., greater than the largest eigenvalue of the coherence matrix $\mathbf{D}^T\mathbf{D}$. In essence, the term $d(\mathbf{x},\mathbf{x}_\theta)$ is a measure of proximity to a previous solution $\mathbf{x}_\theta$. If the vector difference $\mathbf{x}-\mathbf{x}_\theta$ is spanned by $\mathbf{D}$, the distance drops to nearly zero (We usually choose $c=\|\mathbf{D}\|_2^2$). Then, we remain with a minimization over the original cost function $f(\mathbf{x})$. Alternatively, if $\mathbf{D}$ is not full rank and the change $\mathbf{x}-\mathbf{x}_\theta$ is close to the null space of $\mathbf{D}$, the distance is simply the approximate Euclidean distance between the current solution to the previous one.

The surrogate function $Q(\mathbf{x},\mathbf{x}_\theta)$ obeys equality at $\mathbf{x}_\theta$: $Q(\mathbf{x}_\theta,\mathbf{x}_\theta) = f(\mathbf{x}_\theta)$. It is upper-bounded by the original function: $Q(\mathbf{x},\mathbf{x}_\theta) \geq f(\mathbf{x}) \quad \forall \mathbf{x}$, and tangent at $\mathbf{x}_\theta$: $ \nabla Q(\mathbf{x},\mathbf{x}_\theta) |_{\mathbf{x}= \mathbf{x}_\theta} = \nabla f(\mathbf{x}) |_{\mathbf{x}=\mathbf{x}_\theta}$. Hence, the solution sequence is guaranteed to yield decreasing values of the original cost function $f(\mathbf{x})$ because
\begin{flalign*}
f(\mathbf{x}_{\theta+1}) & \leq Q(\mathbf{x}_{\theta+1},\mathbf{x}_{\theta}) = 
\min_{\mathbf{x}} Q(\mathbf{x},\mathbf{x}_{\theta}) \\
& \leq 
Q(\mathbf{x}_{\theta},\mathbf{x}_{\theta}) = f(\mathbf{x}_{\theta}).
\end{flalign*}

Following the MM strategy, that is, minimizing $Q(\mathbf{x},\mathbf{x}_\theta)$ instead of $f(\mathbf{x})$, the sequence of iterative solutions is generated by the recurrent formula
\begin{equation*}
\mathbf{x}_{\theta+1} =  \arg \min_{\mathbf{x}} Q(\mathbf{x},\mathbf{x}_\theta),
\end{equation*}
where $\theta \in \mathbb{N}$ is the iteration index.
Therefore, we can find a closed-form solution for its global minimizer
\begin{equation*}
\mathbf{x}_{\theta+1} =  \mathcal{S}_{\frac{\lambda}{c}}\Big(
\frac{1}{c} \mathbf{D}^T
(\mathbf{y}-\mathbf{D}\mathbf{x}_\theta)
+ \mathbf{x}_\theta \Big).
\end{equation*}
Intuitively, this sequence can be interpreted as an iterative projection of the dictionary on the residual term, starting from the initial solution that is simply a thresholded projection of the dictionary on the observation signal (assuming $\mathbf{x}_0=\mathbf{0}$):
\begin{equation*}
\mathbf{x}_{\theta+1} =  \mathcal{S}_{\frac{\lambda}{c}}\Big(
\frac{1}{c} 
\overbrace{\mathbf{D}^T}^\text{project on dictionary}
\underbrace{(\mathbf{y}-\mathbf{D}\mathbf{x}_\theta)}_\text{residual term} \quad
 + \underbrace{\mathbf{x}_\theta}_\text{add to current solution} \Big).
\end{equation*}

Under the assumption that the constant $c$ is large enough, it was shown in \cite{Daubechies:2004}, that the above algorithm is guaranteed to converge to its global minimum.
Hence, we are guaranteed to recover a local minimum of $f(\mathbf{x})$.
This approach can also be viewed as a proximal-point algorithm \cite{Combettes:2005}, or as a simple projected gradient descent algorithm.

It is worth mentioning that ISTA can be also viewed as a Recurrent Neural Net (RNN) unfolded through time \cite{Papyan:2017}. As previously stated, despite its simplicity, ISTA is considered as a slow algorithm. Over time, faster extensions have been suggested, such as: Fast-ISTA (FISTA) \cite{Beck:2009}, Learned-ISTA (LISTA) \cite{Lecun:2010} and Ada-LISTA \cite{Aberdam:2020}. We refer the reader to the corresponding references for further details.

\subsection{Convolutional Sparse Coding}
In the special case where $\mathbf{D}$ is a convolutional dictionary, the task of extracting $\mathbf{x}$ is referred to as convolutional sparse coding (CSC). In this case, the dictionary $\mathbf{D}$ is a convolutional matrix constructed by shifting a local matrix of $m$ filters in all possible positions. Equivalently, let us assume a convolutional dictionary that is structured as a concatenation of $m$ convolution matrices,
\begin{equation}\label{1.6}
\mathbf{D} = [\mathbf{\mathcal{D}}_1, \mathbf{\mathcal{D}}_2, ... ,\mathbf{\mathcal{D}}_m],
\end{equation}
where $\mathbf{\mathcal{D}}_p \in \mathbb{R}^{L_\mathrm{y} \times L_\mathrm{x} }, \ p\in \mathbb{N}, \ 1 \leq p \leq m$ is the convolution matrix of the $p$'th filter denoted by $\tilde{\mathbf{d}}_p \in \mathbb{R}^{L_\mathrm{d} \times 1}$ shifted in all possible directions. We assume a linear convolution, such that $L_\mathrm{y}=L_\mathrm{x}+L_\mathrm{d}-1$. Accordingly, the sparse weights vector $\mathbf{x} \in \mathbb{R}^{m L\mathrm{x} \times 1}$ obeys
\begin{equation}
\mathbf{x} = [\mathbf{x}^T_{\mathbf{\mathcal{D}}_1}, \mathbf{x}^T_{\mathbf{\mathcal{D}}_2}, ... ,\mathbf{x}^T_{\mathbf{\mathcal{D}}_m}]^T,
\end{equation}
where $L_\mathrm{x}$ is the support size for each of the filters, and $\mathbf{x}_{\mathbf{\mathcal{D}}_p}, \ p\in \mathbb{N}, \ 1 \leq p \leq m $ is the sparse weights vector corresponding to the $p$th filter shifts along the signal. Overall, there are $M=m L_{\mathrm{x}}$ atoms in the dictionary, such that the $i$th atom $\mathbf{d}_i$, $i= (p-1) L_{\mathrm{x}} + l$, $l \in \mathbb{N}, 1 \leq l \leq L_{\mathrm{x}}$ is the $p$th filter's shift to the $l$th support location. 

In the forward-pass of a each layer in a CNN, the input is convolved with a set of learned filters. Then, we apply a pointwise nonlinear function to the computed feature map summed with a bias term. This process can be viewed as equivalent to the layered thresholding algorithm for the CSC model \cite{Papyan:2017}. In other words, the forward pass of a CNN is inherently based on revealing an estimate of a hidden convolutional sparse code of a given signal.

Since CNNs were inspired by the study of the visual brain cortex, the term receptive field is borrowed to describe a small limited local visual area that a neuron reacts to \cite{Geron:2017}. In other words, in CNNs, a neuron located in a certain layer is connected only to the output of neurons in a limited small area of the previous layer. Considering (\ref{1.6}), since the $i$th atom of dictionary $\mathbf{\mathcal{D}}_p$ is the $p$th filter shifted to the the $l$th support location, having a small restricted support of $L_{\mathrm{d}}$ samples around $l$ in the data, each of these small local support areas is referred to as a receptive field.

\subsection{The Signature Dictionary}
The signature dictionary \cite{Aharon:2008} is essentially a convolutional dictionary of a single filter ($m=1$). In this case we attempt to represent a signal solely by one small kernel. An elaborated discussion on the learning of the signature dictionary and its signal representation can be found in \cite{Aharon:2008}.

\section{Problem Formulation} \label{sec3}

In this section, the mathematical formulation focuses on the settings of the seismic inversion task. Nonetheless, the model can be applied to a wide range of applications (such as, medical-imaging, computer-vision, pattern recognition, etc.) and can be incorporated in any system employing CNNs and RNNs. The method is not restricted to signature dictionaries, and can be applied to convolutional dictionaries with more than one filter, and to non-stationary convolutional operators as described below.

\subsection*{Signal model}

Consider an unknown 2D reflectivity signal $\mathbf{X} \in \mathbb{R}^{L_\mathrm{x}  \times J}$ of $J$ channels representing the true reflectivity cross section. We assume a layered subsurface structure and acoustic waves propagation, where reflections are generated at acoustic impedance boundaries. Hence, each hidden column of index $l$ in the reflectivity image is a 1D signal $\mathbf{x}^{(l)}\in \mathbb{R}^{L_\mathrm{x} \times 1}$ independently modeled as a \textit{sparse} weights vector. In a discrete setting, we consider a set of two-way travel-times $T=\{t_m\}$ lying on a grid $k T_\mathrm{s}, \ k\in\mathbb{Z}$, where $t_m=k_m T_\mathrm{s}$, with a sampling rate $F_\mathrm{s}=\frac{1}{T_\mathrm{s}}$, corresponding to a reflector's time-depth location in the ground. Accordingly, a 1D reflectivity signal is formulated as
\begin{equation}\label{2.1}
	x^{(l)}[k]=\sum_m{c_m\delta[k-k_m]}, \quad k \in \mathbb{Z}, \ c_m \in \mathbb{R}, \ l=1,...,J,
\end{equation}
where $\delta[k]$ denotes the Kronecker delta function \cite{Ricker}, $\sum_m|c_m|<\infty$, and
$K=\{k_m\}$ is the set of discrete time delays.

We assume the support $K$ is sufficiently separated. In other words, it obeys the minimal separation condition (see \cite[Definition 2.2]{Pereg:2017A}), with a separation constant $\nu$. Namely,
\begin{equation*}
		\Delta_k \triangleq \min\limits_{k_m,k_n\in K,m\neq n}{\left|k_m-k_n\right|\geq F_\mathrm{s} \nu \sigma}.
\end{equation*} 
where $\sigma>0$ is a given kernel scaling. $\Delta_t \triangleq \nu\sigma$ is the smallest time interval between two reflectors, for which we are guaranteed to perfectly recover two distinct spikes in a noise-free environment. 
In \cite{Pereg:2017A} we prove that under the minimal separation condition, in a noise-free environment, $x^{(l)}[k]$ is perfectly recovered by solving a constrained $\ell_1$ norm optimization problem. We also presented theoretical bounds on the seismic reflectivity recovery error and on the localization error, based on earth Q model.

In a time-variant model we take into account the attenuation and dispersion of the reflected pulses recorded at the geophones on the ground. In this case, $\mathbf{y}^{(l)} \in \mathbb{R}^{L_\mathrm{y} \times 1}$, an observed seismic discrete trace of channel $l$, in the observed seismic 2D data $\mathbf{Y} \in \mathbb{R}^{L_\mathrm{y}  \times J}$, is of the form
\begin{equation}\label{2.2}
	y^{(l)}[k]=\sum_n{x^{(l)}[n]g_{\sigma,n}[k-n]}+w^{(l)}[k], \qquad n \in \mathbb{Z}
\end{equation}
where  $\{g_{\sigma,n}\}$ is a known set of kernels (pulses) corresponding to a possible set of time delays \cite{Pereg:2017A}. 
As stated before, $\sigma>0$ is a known scaling parameter, and $w^{(l)}[k]$ is an additive noise signal. The shape of each pulse $g_{\sigma,n}$ depends on the time (depth) $t_n$ it corresponds to, and the subterrain characteristics, that can be mathematically described by the earth Q model \cite{Pereg:2017A,Wang3,Wang1,Kjartansson}. A brief review of the time-variant pulses model-based estimation can be found in Appendix D. 

Alternatively, one can assume a conventional convolution model where the wavelet is time-invariant.
In other words, all kernels are identical: $g_{\sigma,n}[k]=g_\sigma[k]\ \forall n$.
Namely, in the time-invariant case, each seismic observed trace can be described as
\begin{equation}\label{2.3}
	y^{(l)}[k]=\sum_n{x^{(l)}[n]g[k-n]}+w^{(l)}[k], \qquad n \in \mathbb{Z}.
\end{equation}
where  $g[k]$ is a seismic wavelet of length $L_\mathrm{g}$, and $w^{(l)}[k]$ is an additive noise. Clearly $L_\mathrm{y}=L_\mathrm{x}+L_\mathrm{g}-1$. The wavelet is assumed to be invariant in both time and space (i.e., both in horizontal and vertical directions). We assume that the seismic signal is free of multiple reflections \cite{Sheriff}.


In matrix-vector form we can model the observed 2D seismic data image $\mathbf{Y} \in \mathbb{R}^{L_\mathrm{y}\times J}$ in both cases as
\begin{equation}\label{2.4}
			\mathbf{Y}=\mathbf{GX}+\mathbf{W}.
\end{equation}
Generally, $\mathbf{G}$ is an operator matrix such that $G_{k,n}=g_{\sigma,n}[k-n]$.
In the time-invariant case $\mathbf{G}$ is a convolution matrix of size $L_\mathrm{y} \times L_\mathrm{x}$ such that $G_{k,n}=g[k-n]$, and $\mathbf{W}$ is an additive i.i.d white noise matrix independent of $\mathbf{X}$, with zero mean and variance $\sigma^2_w$. We do not impose any prior knowledge of the structure or possible patterns in the reflectivity image.

Note that practically, even in a noise-free scenario under the separation condition, $\mathbf{x}^{(l)}$ and $\mathbf{G}$ do not obey the bound guaranteeing neither a unique solution for $P_1$ nor a stable solution for $P_{1,\varepsilon}$. Namely, in most practical cases we have,
\begin{equation*}
			\| \mathbf{x}^{(l)} \|_0 > \frac{1}{2}\Big( 1+ \frac{1}{\mu(\mathbf{G})} \Big).
\end{equation*}
The mathematical analysis takes a worst-case point of view. For this reason, the stability and success guarantees are known to be quite pessimistic. Tighter bounds could have probably been obtained. Yet, this bound is still widely used, perhaps due to its simplicity.

\section{Proposed Method}\label{sec4}

As described in Section~\ref{sec2}, one way to look at the seismic inversion problem is to simply solve a CSC problem, trace by trace, using one of the popular methods (such as ISTA, FISTA, LISTA, ada-LISTA, etc. \cite{Daubechies:2004,Beck:2009,Lecun:2010,Aberdam:2020}). Classical pursuit methods, such as ISTA, for example, are guaranteed to recover the true unique solution to the $P_0$ problem \cite{Elad} providing that the number of non-zeros per stripe is less than $\frac{1}{2}\Big(1+\frac{1}{\mu(\mathbf{D})}\Big)$. In real-life scenarios, where we usually consider the $P_{1,\varepsilon}$ problem, stability of the results is also guaranteed under the assumption that $\mathbf{x}$ is sparse enough. Namely, 
if $\|\mathbf{x}\|_0 < \frac{1}{2}\Big(1+\frac{1}{\mu(\mathbf{D})}\Big)$, than the deviation of the recovery $\hat{\mathbf{x}}$ from the true $\mathbf{x}$ is bounded by \cite{Elad}
\begin{equation*}
\|\mathbf{x}-\hat{\mathbf{x}}\|_2^2 \leq \frac{4\varepsilon^2}{1-\mu(\mathbf{D})(2\|\mathbf{x}\|_0-1)}.
\end{equation*}

Nevertheless, in many practical cases, we observe four major issues inherent to the data:
\begin{enumerate}
\item The mutual coherence of the dictionary is relatively high. For example, in the seismic scenario usually $0.5 <\mu(\mathbf{G}) < 1$ (depending on the sampling rate and the wavelet's scaling $\sigma$ and on the $Q$ attenuation factor).
\item The signal (or even a signal stripe (or a patch)) is not sufficiently sparse for a successful recovery (see \cite{Papyan:2017}).
\item The ratio between the global maximal and minimal $\mathbf{x}$ values in absolute value $\frac{|\mathbf{x}|_{\mathrm{max}}}{|\mathbf{x}|_{\mathrm{min}}}$ is high. This in turn leads to longer convergence time and erroneous results due to difficulty to determine the required parameters, especially the thresholds (biases).
\item Recovering the sparse code takes too much time and cannot be applied to real-time applications.
\end{enumerate}

Motivated by these challenges, we suggest a simple modification in the conventional approach.
Besides its simplicity, the main advantage of the proposed method is a substantial speed up, without requiring any pre-training. Our approach could also be potentially incorporated in conventional neural nets forward pass.

\subsection{Receptive Field Normalization Iterative Thresholding Algorithm}

Most thresholding algorithms are inherently limited by the challenge of setting a (global or local) threshold that is not inclined towards spikes of strong amplitudes, even when assuming sufficient separation, and imposing low mutual coherence. In other words, when projecting the signal on the dictionary, in order to detect the presence of a dictionary atom in the signal (or the residual), we need to choose a threshold that would fit strong spikes as well as small spikes. A threshold that is too small would yield a smeared solution (not sparse enough), whereas a threshold that is too large results in missed spikes (too sparse). To cope with this problem ISTA \cite{Daubechies:2004} repeatedly iterates over the residual term, with a constant yet relatively small threshold, which makes convergence considerably slow. In ISTA, the threshold $\frac{\lambda}{c}$ controls the desired sparsity. To cope with this issue, FISTA incorporates a momentum term in the update step at each iteration \cite{Beck:2009}. On the other hand, LISTA \cite{Lecun:2010} and Ada-LISTA \cite{Aberdam:2020} attempt to learn more suitable weight matrices and use adaptive thresholds.

We propose a different approach. That is, instead of modifying the threshold and/or the dictionary, or having learned different local thresholds for each receptive field (as can be done in CNNs), we normalize the energy of each receptive field, before projecting it on the features space.

\textit{Definition 1: Receptive Field Normalization Kernel}\\
A kernel $h[k]$ can be referred to as a receptive field normalization kernel if 
\begin{enumerate}
\item The kernel is positive: $h[k] \geq 0 \quad \forall k$.
\item The kernel is symmetric: $h[k]=h[-k] \quad \forall k$.
\item The kernel's global maximum is at its center: $h[0] = 1 \geq h[k] \quad \forall k \neq 0$.
\item The kernel's energy is finite: $ \sum_k {h[k]} < \infty$.
\end{enumerate}

\textit{Definition 2: Receptive Field Normalization}\\
Assuming a receptive field normalization kernel $h[k]$ of odd length $L_\mathrm{h}$, we define the local weighted energy of a time window centered around the $k$th sample of a 1D observed data signal $\mathbf{y}\in\mathbb{R}^{L_\mathrm{y} \times 1}$
\begin{equation}\label{3.1}
\sigma_{\mathrm{y}}[k] \triangleq \Bigg( \sum_{n=-\frac{L_\mathrm{h}-1}{2}}^{\frac{L_\mathrm{h}-1}{2}} { h[n] y^2[k-n] } \Bigg) ^ \frac{1}{2}.
\end{equation} 
When $y[k]$ is modeled in accordance to (\ref{2.2}), or any other application where $\mathbf{y}=\mathbf{Dx}+\mathbf{e}$, and $h[k]$ is a receptive field normalization window function of length $L_\mathrm{h} \leq L_\mathrm{d}$ odd number of samples. For our application we used a truncated Gaussian-shaped window, but it is possible to use any other window function depending on the application, such as: a rectangular window, Epanechnikov window, etc. The choice of the normalization window and its length affects the choice of the thresholding parameters. If $h[k]$ is a rectangular window, then $\sigma_\mathrm{y}[k]$ is simply the $\ell_2$ norm of a data stripe centered around the $k$th location. Otherwise, if the chosen receptive field normalization window is attenuating, then the energy is focused in the center of the receptive field, and possible events at the margins are repressed. The window size is $L_\mathrm{h} \leq L_\mathrm{d}$ so as to avoid interference of adjacent events as much as possible. Yet it is recommended not to use a winodw that is too small. $L_\mathrm{h} \geq \frac{L_\mathrm{d}}{2}$ can serve as a good rule of thumb.

Receptive field normalization is employed by dividing each data point by a local energy measure, before projecting the signal on the dictionary and taking a threshold. Assuming a signal $\mathbf{y}$ and a convolutional dictionary $\mathbf{D}$, an initial solution $\mathbf{x}_0=\mathbf{0}$ and an initial estimated support $\mathbf{\Delta q}_0=\mathbf{0}$ at $\theta=0$. At the first iteration $\theta=1$, we compute the local variance of $y[k]$ as defined in (\ref{3.1}), namely,
\begin{equation}\label{3.2}
\sigma_\mathrm{y}[k]=\sqrt{h[k]*y^2[k]},
\end{equation}
where $h[k]$ is a receptive field normalization window, and $*$ denotes the convolution operation.
Then, we normalize the signal by dividing each data point by the corresponding receptive field energy.
In order to avoid amplification of low energy regions, we use a clipped version of $\sigma_\mathrm{y}[k]$. Namely,
\begin{equation}\label{3.3}
\tilde{\sigma}_\mathrm{y}[k]=
\begin{cases}
\sigma_\mathrm{y}[k] &  \big| \sigma_\mathrm{y}[k] \big| \geq \tau_1 \\
1 & \big| \sigma_\mathrm{y}[k] \big| < \tau_1
\end{cases},
\end{equation}
where $\tau_1>0$ is a predetermined threshold for the first iteration. Empirically, for our application $0.15 \leq \tau_1 \leq 0.4$ works well. 
We define the signal normalization weight matrix $\mathbf{W}_0 \in \mathbb{R}^{L_{\mathrm{y}} \times L_{\mathrm{y}}}$,
\begin{equation}\label{3.4a}
\mathbf{W}_0=\mathrm{diag} \Bigg(\frac{1}{\tilde{\sigma}_\mathrm{y}[k]} \Bigg), \  k=1,...,L_{\mathrm{y}}.
\end{equation}
Also, we define the dictionary normalization weight matrix as
\begin{equation}\label{3.4b}
\mathbf{W}_\mathrm{D}=\mathrm{diag} \Bigg(\frac{1}{\sigma_{\mathrm{d}_i}} \Bigg), \ i=1,...,m L_{\mathrm{x}}.
\end{equation}
where $\sigma_{\mathrm{d}_i}=\|\mathbf{d}_i\|_2$ is the $\ell_2$ norm of the $i$th atom. Recall that since the dictionary is convolutional, the $i$th atom, $i=(k-1)L_\mathrm{x}+l$ is the $k$th filter shifted to the $l$th support location. Clearly, when the dictionary is a time-invariant convolutional dictionary, $\sigma_{\mathrm{d}_{(k-1)L_\mathrm{x}+l}}=\sigma_{\mathrm{d}_k} \ \forall k \in [1,m]$.
  
The detected support at the first iteration is
\begin{equation}\label{3.5}
\mathbf{q}_1=\mathcal{I}_{\beta_1}(\mathbf{W}_\mathrm{D} \mathbf{D}^T\mathbf{W}_0\mathbf{y}),
\end{equation}
where $\mathcal{I}$ is an element-wise thresholding indicator function
\begin{equation}
\mathcal{I}_{\beta_1}(x_k)=
\begin{cases}
1 &  |x_k| \geq \beta_1 \\
0 &  |x_k| < \beta_1.
\end{cases}
\end{equation}
If the dictionary atoms are assumed to be normalized, such that $\sigma_{\mathrm{d}_i}=1  \ \forall i$, then of course $\mathbf{W}_\mathrm{D}$ is simply an identity matrix, and one can simply ignore $\mathbf{W}_\mathrm{D}$ throughout the entire formulation.

It is a well known fact that according to Cauchy-Schwartz inequality the correlation coefficient defined as 
\begin{equation}\label{3.6}
\rho_{ab} \triangleq \frac{\mathbf{a}^T\mathbf{b}}{\|\mathbf{a}\|_2 \|\mathbf{b}\|_2}
\end{equation}
is bounded by one (in absolute value): $|\rho_{\mathrm{ab}}| \leq 1$. Therefore, if we were to divide the inner product $\mathbf{D}^T\mathbf{y}$ by each receptive field's energy and by the corresponding atom's energy, we could detect an atom's sole presence and a perfect match in a time-window of the signal $\mathbf{y}$, when the result is exactly 1. 
Of course, when several pulses interfere in a single time window, when the mutual coherence is not small enough, the threshold needs to be adjusted accordingly, and a few iterations for corrections may be needed. Choosing an attenuating normalization kernel can repress adjacent spikes. Overall, accurate detection depends on the correlation between neighboring atoms and on neighboring spikes amplitudes, as analytically described in Theorems~\ref{Theorem 1}-\ref{Theorem 3}.

As stated below, if the support is sufficiently separated, and the mutual coherence is sufficiently small, then the true support is \textit{perfectly} recovered at this stage, within the first iteration. Once the support is recovered there are two simple ways to determine the amplitudes. The first, and more accurate, is simply to solve an LS problem for the subsystem $\mathbf{y}=\mathbf{D}_{K_1}\mathbf{x}_{K_1}$, where $\mathbf{D}_{K_1}$ denotes the partial dictionary matrix having $|{K_1}|$ columns from the columns of $\mathbf{D}$ with indices in ${K_1}$  - the estimated support in the first iteration. Alternatively, if one wishes to speed-up the algorithm, or in real-life applications where the support is not sufficiently sparse, it is possible to approximate the amplitudes simply as $\mathbf{x}_1=\mathbf{q}_1 \odot \big(\mathbf{W}^2_\mathrm{D}\mathbf{D}^T \mathbf{y}\big)$, where $\odot$ denotes the Hadamard product.

In the special case of a signature dictionary ($m=1$), it is possible to approximate the support amplitudes by $x[k]=y[k+\Delta_d]/d_{k}^p$, where $k \in K_1$ is a support index, $\Delta_d \triangleq \frac{L_\mathrm{d}-1}{2}$ is the convolution time shift, and $d_{k}^p=d_{k}[k+\Delta_d]$ denotes the corresponding atom's central value. For a signature dictionary of a Ricker wavelet, $d_{k}^p= \|\mathbf{d}_k \|_{\infty}=g[0]=1, \forall k$. Empirically, we have not witnessed a significant advantage for the first accurate method over the approximate one, in the case of the signature dictionary. For the sake of brevity, in this subsection, we assume without loss of generality that the dictionary atoms central value is one, i.e., $d_{k}^p=1, \forall k$.

If the support is sufficiently separated, the algorithm is done at this stage, within one iteration. Otherwise, we can proceed to perform iterative stages as follows. At each iteration, we find the required change in the support by projecting the dictionary on the normalized residual.
The residual at iteration $\theta$ is 
\begin{equation}\label{3.7}
	\mathbf{\Delta r}_{\theta+1}=\mathbf{y-Dx}_\theta.
\end{equation} 
We now compute the local weighted variance of the residual term
\begin{equation}\label{3.8}
\sigma^{\theta+1}_{\Delta r}[k]=\sqrt{h[k]* \Delta r_{\theta+1}^2[k]},
\end{equation}
and its clipped version
\begin{equation}\label{3.9}
\tilde{\sigma}^{\theta+1}_{\Delta r}[k]=
\begin{cases}
\sigma^{\theta+1}_{\Delta r}[k] & \big| \sigma^{\theta+1}_{\Delta r}[k] \big| \geq \tau_\theta \\
1 & \big| \sigma^{\theta+1}_{\Delta r}[k] \big| < \tau_\theta
\end{cases},
\end{equation}
Once again, we build the corresponding normalization weight matrix,
\begin{equation}\label{3.10}
\mathbf{W}_{\theta}=\mathrm{diag} \Bigg( \frac{1}{\tilde{\sigma}^{\theta+1}_{\Delta r}[k]} \Bigg), \ k=1,...,L_{\mathrm{y}}.
\end{equation}
Then, we project the dictionary on the normalized residual and threshold the obtained signal,
\begin{flalign}\label{3.11}
\nonumber
\Delta \mathbf{q}_{\theta+1} 
& =\mathcal{I}_{\beta_{\theta+1}}(\mathbf{W}_\mathrm{D} \mathbf{D}^T\mathbf{W}_\theta	\mathbf{\Delta r}_{\theta+1}) 
\\
& =\mathcal{I}_{\beta_{\theta+1}}( \mathbf{W}_\mathrm{D} \mathbf{D}^T\mathbf{W}_{\theta}(\mathbf{y-Dx}_\theta)).
\end{flalign}
The updated solution at iteration $\theta+1$ is
\begin{equation}\label{3.12}
	\mathbf{x}_{\theta+1}=\Delta \mathbf{q}_{\theta+1} \odot \Big(\mathbf{W}^2_\mathrm{D} \mathbf{D}^T \Delta \mathbf{r}_{\theta+1}\Big) + \mathbf{x}_{\theta},
\end{equation}
where $\odot$ denotes the Hadamard product.
Note that the algorithm should converge in a few iterations. About 2-5 iterations should be enough. To ensure stable convergence, incorporating a momentum, is also possible:
\begin{equation}\label{3.13a}
	\mathbf{x}_{\theta+1}= \alpha_\mathrm{r} \Big( \Delta \mathbf{q}_{\theta+1}\odot \Big(\mathbf{W}^2_\mathrm{D} \mathbf{D}^T \Delta \mathbf{r}_{\theta+1}\Big)  \Big) +\mathbf{x}_{\theta},
\end{equation}
where $0 < \alpha_\mathrm{r} \leq 1$ is a constant step size.
In practical cases, when the data is not sufficiently separated and the mutual coherence is relatively high, an exact stopping rule may be hard to determine and solutions may be unstable. In these cases, early stopping after a predetermined small number of iterations is recommended. Note that as opposed to ISTA, here the shrinkage operator is applied only to the projection on the residual, without formerly adding it to the previous solution.
As previously stated, in the case of a signature dictionary it is possible to use
\begin{equation}\label{3.13b}
	\mathbf{x}_{\theta+1}= \alpha_\mathrm{r} \Big( \Delta \mathbf{q}_{\theta+1}\odot \Delta \mathbf{r}_{\theta+1}  \Big) +\mathbf{x}_{\theta},
\end{equation}
instead of (\ref{3.13a}).

From a neural network perspective, receptive-field normalization (RFN) increases the sensitivity of a neuron to its receptive field. In some cases, where the energy of receptive fields in different locations is unbalanced, some atoms may be strongly expressed, whereas others are significantly weaker. RFN overcomes this obstacle, without having to set local or adaptive thresholds. This way, the activation of a neuron is independent of the scaling of other events in the signal. In other words, the neuron is able to detect a feature (reflection), even if its energy is relatively low, comparing to other events in the data.  

Note that $\tau_\theta$ essentially determines the minimal $|\mathbf{x}|_{\mathrm{min}}$ that can be detected at each of the above steps. It should be determined taking into account the noise expected level. Namely, since we usually assume the nuisance noise is uncorrelated with the signal, therefore,
\begin{equation*} 
E \|\mathbf{y}\|^2_2   = E  \|\mathbf{Dx}\|^2_2  + E\|\mathbf{e}\|^2_2 ,
\end{equation*}
where $E$ denotes mathematical expectation.
In order to avoid noise amplification, we would set $\tau_\theta$ such that
\begin{equation} \label{3.14}
\tau_\theta \geq  |\mathbf{x}|_{\mathrm{min}} \min_i \|\mathbf{d}_i\|_2 + \varepsilon_\mathrm{d},
\end{equation}
where $\varepsilon_{\mathrm{d}}$ denotes the noise $\ell_2$ norm over a data stripe of length $L_\mathrm{h}$.

We project the dictionary on the normalized signal though it is possible to normalize the projection $\mathbf{D}^T(\mathbf{y}-\mathbf{Dx}_\theta)$ instead. Yet, the two options are not equivalent. Normalization of the signal prior to projection as in (\ref{3.10}) rescales each sample, and therefore might cause some distortion to the signal. It is more suitable for admissible kernels (see \cite[Definition 2.1]{Pereg:2017A}), where most of the kernel's energy is focused at its center. On the other hand, it promotes muting of close spikes and decreases noise influence especially when dealing with relatively small environments.

A summary of the proposed methods is presented in Algorithms \ref{alg1}-\ref{alg3}.

\SetKwInput{kwInit}{Init}
\begin{algorithm}[t]\label{alg1}
\SetAlgoLined
\SetKwInOut{Input}{input}\SetKwInOut{Output}{output}
\Input{signal $\mathbf{y}$, dictionary $\mathbf{D}$}
\kwInit{$\mathbf{x}_0=\mathbf{0}$, $\mathbf{\Delta q}_0=\mathbf{0}$, $\theta=0$} 
compute: $\mathbf{W}_\mathrm{D}=\mathrm{diag} \Big(\sigma_{\mathrm{d}_i}^{-1} \Big), i=1,...,m L_{\mathrm{x}}.$\\

 \While{$\|\mathbf{x}_{\theta+1}-\mathbf{x}_\theta\|_2< \delta $ or $\theta \leq N_{\mathrm{it}}$}{
	$\mathbf{\Delta r}_{\theta+1}=\mathbf{y-Dx}_\theta$\\
	compute:
	\begin{equation*}
	\tilde{\sigma}^{\theta+1}_{\Delta r}[k]=
	\begin{cases}
	\sigma^{\theta+1}_{\Delta r}[k] &  \left|\sigma^{\theta+1}_{\Delta r}[k] \right| \geq \tau_\theta \\
	1 & \left|\sigma^{\theta+1}_{\Delta r}[k] \right| < \tau_\theta
	\end{cases},
	\end{equation*}\\
	$\mathbf{W}_{\theta}=\mathrm{diag} \Big(\tilde{\sigma}^{\theta+1}_{\Delta r}[k] \Big)^{-1}, k=1,...,L_{\mathrm{y}}$\\
  $\Delta \mathbf{q}_{\theta+1}=\mathcal{I}_{\beta_{\theta+1}}(\mathbf{W}_\mathrm{D} \mathbf{D}^T\mathbf{W}_\theta	\mathbf{\Delta r}_{\theta+1})$\\
	solve the subsystem $\mathbf{\Delta r}_{\theta+1}=\mathbf{D}_{K_{\theta+1}}\Delta \mathbf{x}_{\theta+1}$,\\ 
	$\mathbf{D}_{K_{\theta+1}}$ - partial dictionary matrix corresponding to the support $\Delta \mathbf{q}_{\theta+1}$.\\
	$\mathbf{x}_{\theta+1}= \alpha_\mathrm{r} \Delta \mathbf{x}_{\theta+1}+\mathbf{x}_{\theta}$\\
	$\theta \leftarrow \theta+1$ \\
 }
 \caption{Receptive Field Normalization Iterative Thresholding Algorithm}
\end{algorithm}

\SetKwInput{kwInit}{Init}
\begin{algorithm}[t]\label{alg2}
\SetAlgoLined
\SetKwInOut{Input}{input}\SetKwInOut{Output}{output}
\Input{signal $\mathbf{y}$, dictionary $\mathbf{D}$}
\kwInit{$\mathbf{x}_0=\mathbf{0}$, $\mathbf{\Delta q}_0=\mathbf{0}$, $\theta=0$} 
compute: $\mathbf{W}_\mathrm{D}=\mathrm{diag} \Big(\sigma_{\mathrm{d}_i}^{-1} \Big), i=1,...,m L_{\mathrm{x}}.$\\
 \While{$\|\mathbf{x}_{\theta+1}-\mathbf{x}_\theta\|_2< \delta $ or $\theta \leq N_{\mathrm{it}}$}{
	$\mathbf{\Delta r}_{\theta+1}=\mathbf{y-Dx}_\theta$\\
	compute:
	\begin{equation*}
	\tilde{\sigma}^{\theta+1}_{\Delta r}[k]=
	\begin{cases}
	\sigma^{\theta+1}_{\Delta r}[k] &  \left|\sigma^{\theta+1}_{\Delta r}[k] \right| \geq \tau_\theta \\
	1 & \left|\sigma^{\theta+1}_{\Delta r}[k] \right| < \tau_\theta
	\end{cases},
	\end{equation*}\\
	$\mathbf{W}_{\theta}=\mathrm{diag} \Big(\tilde{\sigma}^{\theta+1}_{\Delta r}[k] \Big)^{-1}, k=1,...,L_{\mathrm{y}}$\\
  $\Delta \mathbf{q}_{\theta+1}=\mathcal{I}_{\beta_{\theta+1}}(\mathbf{W}_\mathrm{D} \mathbf{D}^T\mathbf{W}_\theta	\mathbf{\Delta r}_{\theta+1})$\\
	$\mathbf{x}_{\theta+1}= \alpha_\mathrm{r} \Big( \Delta \mathbf{q}_{\theta+1} \odot \Big( \mathbf{W}^2_\mathrm{D}\mathbf{D}^T \Delta \mathbf{r}_{\theta+1} \Big) \Big)+\mathbf{x}_{\theta}$\\
	$\theta \leftarrow \theta+1$ \\
 }
 \caption{Approximate Receptive Field Normalization Iterative Thresholding Algorithm}
\end{algorithm}

\SetKwInput{kwInit}{Init}
\begin{algorithm}[t]\label{alg3}
\SetAlgoLined
\SetKwInOut{Input}{input}\SetKwInOut{Output}{output}
\Input{signal $\mathbf{y}$, dictionary $\mathbf{D}$}
\kwInit{$\mathbf{x}_0=\mathbf{0}$, $\mathbf{\Delta q}_0=\mathbf{0}$, $\theta=0$} 
compute: $\mathbf{W}_\mathrm{D}=\mathrm{diag} \Big(\sigma_{\mathrm{d}_i}^{-1} \Big), i=1,...,m L_{\mathrm{x}}.$\\
 \While{$\|\mathbf{x}_{\theta+1}-\mathbf{x}_\theta\|_2< \delta $ or $\theta \leq N_{\mathrm{it}}$}{
	$\mathbf{\Delta r}_{\theta+1}=\mathbf{y-Dx}_\theta$\\
	compute:
	\begin{equation*}
	\tilde{\sigma}^{\theta+1}_{\Delta r}[k]=
	\begin{cases}
	\sigma^{\theta+1}_{\Delta r}[k] &  \left|\sigma^{\theta+1}_{\Delta r}[k] \right| \geq \tau_\theta \\
	1 & \left|\sigma^{\theta+1}_{\Delta r}[k] \right| < \tau_\theta
	\end{cases},
	\end{equation*}\\
	$\mathbf{W}_{\theta}=\mathrm{diag} \Big(\tilde{\sigma}^{\theta+1}_{\Delta r}[k] \Big)^{-1}, k=1,...,L_{\mathrm{y}}$\\
  $\Delta \mathbf{q}_{\theta+1}=\mathcal{I}_{\beta_{\theta+1}}(\mathbf{W}_\mathrm{D} \mathbf{D}^T\mathbf{W}_\theta	\mathbf{\Delta r}_{\theta+1})$\\
	$\mathbf{x}_{\theta+1}= \alpha_\mathrm{r} \Big( \Delta \mathbf{q}_{\theta+1} \odot \Delta \mathbf{r}_{\theta+1} \Big)+\mathbf{x}_{\theta}$\\
	$\theta \leftarrow \theta+1$ \\
 }
 \caption{Approximate Receptive Field Normalization Iterative Thresholding Algorithm for a Signature Dictionary}
\end{algorithm}

Algorithm \ref{alg4} shows a further simplification of the proposed method. Namely, after computing the locally normalized signal $\tilde{\mathbf{y}} \triangleq \mathbf{W}_0\mathbf{y}$ at the first iteration, we propagate through the iterations without computing the spikes amplitude and without normalizing again, in order to estimate only the support. When the support is fully revealed, we calculate the weights by solving an LS problem or an approximation as described above. This variation can be used if more speed is necessary, on the expense of accuracy, or for implementing a learned version as discussed is Section~\ref{sec7}.

\SetKwInput{kwInit}{Init}
\begin{algorithm}\label{alg4}
\SetAlgoLined
\SetKwInOut{Input}{input}\SetKwInOut{Output}{output}
\Input{signal $\mathbf{y}$, dictionary $\mathbf{D}$}
\kwInit{$\mathbf{x}_0=\mathbf{0}$, $\mathbf{\Delta q}_0=\mathbf{0}$, $\mathbf{q}_0=\mathbf{0}$, $\theta=0$} 
compute:
\begin{equation*}
\tilde{\sigma}_y[k]=
\begin{cases}
\sigma_y[k] &  |\sigma_\mathrm{y}[k]| \geq \tau_1 \\
1 & |\sigma_\mathrm{y}[k]| < \tau_1
\end{cases},
\end{equation*}

$\mathbf{W}_\mathrm{D}=\mathrm{diag} \Big(\sigma_{\mathrm{d}_i}^{-1} \Big), \ i=1,...,m L_{\mathrm{x}}.$\\
	
$\mathbf{W}_0=\mathrm{diag} \Big(\tilde{\sigma}^{-1}_\mathrm{y}[k] \Big), \ k=1,...,L_{\mathrm{y}}$,\\

$\tilde{\mathbf{y}}= \mathbf{W}_0	\mathbf{y}$\\

 \While{$\|\mathbf{\Delta} \tilde{\mathbf{y}}_{\theta+1}-\mathbf{\Delta} \tilde{\mathbf{y}}_\theta\|_2< \delta $ or $\theta \leq N_{\mathrm{it}}$}{
	$\mathbf{\Delta} \tilde{\mathbf{y}}_{\theta+1}=\tilde{\mathbf{y}}-\mathbf{Dq}_\theta$\\
	
  $\mathbf{\Delta q}_{\theta+1}= \mathcal{I}_{\beta_{\theta+1}}(\mathbf{W}_\mathrm{D} \mathbf{D}^T\mathbf{\Delta \tilde{y}}_{\theta+1})$\\
	$\mathbf{q}_{\theta+1}= \alpha_\mathrm{r} \mathbf{ \Delta q}_{\theta+1}+\mathbf{q}_{\theta}$\\
	$\theta \leftarrow \theta+1$ \\
 }

 $\hat{\mathbf{x}}= \hat{\mathbf{q}} \odot \mathbf{y} $
 \caption{Support Detection Approximate Receptive Field Normalization Iterative Thresholding Algorithm for a Signature Dictionary}
\end{algorithm}
\vspace{-0.5cm}

\subsection{Theoretical Analysis for the RFN-thresholding with a Convolutional Sparse Model}

Consider a stripe of length $L_\mathrm{h}$ around some index $i$ and the corresponding atoms shifts around this location in the signal. Hereafter, the $i$th entry of a vector $\mathbf{v}$ is denoted by $\mathbf{v}[i]$, and a stripe cropped around the $i$th index is denoted by $\mathbf{v}_i$. Note that due to the convolution properties of the dictionary, a data stripe $\mathbf{y}_i \in \mathbb{R}^{L_\mathrm{h} \times 1}$ is affected from input spikes in $\mathbf{x}_i \in \mathbb{R}^{m L_\mathrm{s} \times 1}$, such that $L_\mathrm{s}=L_\mathrm{h}+L_\mathrm{d}-1$. To shed some light on the theoretical aspects of the proposed method, let us define the local sparsity of a stripe of $\mathbf{x}$ as the maximum number of non zeros weights corresponding to a data stripe of length $L_\mathrm{h}$ at location $i$,
\begin{equation}
s = \|\mathbf{x}\|^{S^i_h}_{0,\infty}= \max_i \| \mathbf{x}_i \|_0= \max_i \sum_{l \in S^i_h} \mathcal{I}_0(\mathbf{x}[l]),
\end{equation}
where $S^i_h$ denotes a neighborhood of $m L_\mathrm{s}$ point values corresponding to $L_\mathrm{h}$ data points symmetrically distributed around index $i$, and redefining $\mathcal{I}_0(0)=0$.
Let us rephrase equation (\ref{3.1}) in matrix-vector form
\begin{equation}\label{3.25}
\sigma_{\mathrm{y}}[i] = \| \mathbf{H} \mathbf{y}_i \|_2.
\end{equation} 
where $\mathbf{y}_i$  denotes a stripe of length $L_\mathrm{h}$ around some index $i \in [1, L_{\mathrm{y}}]$, that is,
$\mathbf{y}_i = \big[ y[i-\frac{L_\mathrm{h}-1}{2}],..., y[i + \frac{L_\mathrm{h}-1}{2}] \big]^T$ and $\mathbf{H} = \mathrm{diag}(h^{\frac{1}{2}})$ (ignoring boundary issues). 
Throughout the proofs we assume $\sigma_{\mathrm{y}}[i]>\tau \ \forall i\in[1,L_{\mathrm{y}}]$. 

\begin{theorem}[Support Recovery using RFN in the Presence of Noise]
\label{Theorem 1}

Let $\mathbf{y}=\mathbf{Dx}+\mathbf{e}$, where $\mathbf{D}$ is a convolutional dictionary, with normalized atom's variance $\| \mathbf{d}_i \|_2=\sigma_\mathbf{\mathrm{d}} = 1\ \forall i \in [1, mL_\mathrm{x}]$. Assume a rectangular receptive field normalization kernel, that is $\mathbf{h}[k]=1$, $\mathbf{h} \in \mathbb{R}^{L_{\mathrm{h}}\times 1}$, such that $L_{\mathrm{h}}=L_{\mathrm{d}}$. In other words $\mathbf{H}$ simply extracts a local data stripe.
Assuming that
\begin{enumerate}[wide, labelwidth=!, labelindent=0pt]
\item
\begin{flalign}\label{3.15}
\nonumber
\hspace{-3pt}
 \frac{|\mathbf{x}_i|_\mathrm{min}}{|\mathbf{x}_i|_\mathrm{max}+\frac{\varepsilon_{\mathrm{d}}}{s}} >
\frac{s\mu}{1+\mu} \Bigg( 1 +\frac{\sqrt{s}}{\sqrt{1 -(s-1)\mu}-\tilde{\varepsilon}_{\mathrm{d},\infty} } \Bigg) + \frac{2 s \varepsilon_\mathrm{s}}{1+\mu} \\
\quad \forall i \in K.
\end{flalign}
where $|\mathbf{x}_i|_{\mathrm{max}}$ and $|\mathbf{x}_i|_{\mathrm{min}}$ are the highest and the lowest entries in a stripe $\mathbf{x}_i$ respectively, $\varepsilon_\mathrm{d} \triangleq \| \mathbf{e}_i\|_2$, $\tilde{\varepsilon}_{\mathrm{d},\infty} \triangleq \frac{\varepsilon_{\mathrm{d}}}{|\mathbf{x}|_\mathrm{min}}$ and $\varepsilon_\mathrm{s} \triangleq \frac{\varepsilon_{\mathrm{d}}}{\tau}$.
\item
The threshold $\beta_1$ is chosen according to
\begin{equation}\label{3.16}
\frac{(1+\mu)|\mathbf{x}_i|_\mathrm{min}}{s|\mathbf{x}_i|_\mathrm{max}+\varepsilon_{\mathrm{d}}} - \mu - \varepsilon_\mathrm{s}
>\beta_1>
\frac{\sqrt{s} \mu}{\sqrt{1 -(s-1)\mu} -\tilde{\varepsilon}_{\mathrm{d},\infty} } + \varepsilon_\mathrm{s}.
\end{equation}
\end{enumerate}
Then the support of $\mathbf{x}$ is perfectly recovered with receptive-field normalization thresholding,
\begin{equation*}
\mathrm{Supp}(\mathbf{x}_1)=\mathrm{Supp}(\mathbf{x}),
\end{equation*}
where $\mathrm{Supp}(\cdot)$ is the support of a vector, and $\mathbf{x}_1$ is the recovered sparse vector within one step of RFN thresholding (or one iteration of Algorithms 1-4).
\end{theorem}
\textit{Proof:} see Appendix A.

Consequently, in the noise-free model $\mathbf{y}=\mathbf{Dx}$, the true support is perfectly recovered within one iteration of Algorithms \ref{alg1}-\ref{alg4} provided that
\begin{enumerate}
\item
\begin{equation}\label{3.17}
\frac{|\mathbf{x}_i|_\mathrm{min}}{|\mathbf{x}_i|_\mathrm{max}}  > \frac{s\mu}{1+\mu} \Bigg( 1 +\frac{\sqrt{s}}{\sqrt{1 -(s-1)\mu}} \Bigg) \quad \forall i \in K.
\end{equation}
where $|\mathbf{x}_i|_{\mathrm{max}}$ and $|\mathbf{x}_i|_{\mathrm{min}}$ are the highest anf the lowest entries in a stripe $\mathbf{x}_i$ respectively.
\item The global threshold $\beta_1$ is chosen according to
\begin{equation}\label{3.18}
\frac{1+\mu}{s} \Bigg( \min_{i\in K} \frac{|\mathbf{x}_i|_\mathrm{min}}{|\mathbf{x}_i|_\mathrm{max}}\Bigg) - \mu  
>\beta_1>
\frac{\sqrt{s} \mu}{\sqrt{1 -(s-1)\mu}}.
\end{equation}
\end{enumerate}

The conditions (\ref{3.15}) and (\ref{3.17}) may appear too strict at first glance, but note that we are considering the \textit{local} ratio $\frac{|\mathbf{x}_i|_\mathrm{min}}{|\mathbf{x}_i|_\mathrm{max}}$ rather than the \textit{global} ratio $\frac{|\mathbf{x}|_\mathrm{min}}{|\mathbf{x}|_\mathrm{max}}$. Also the bound is far from being tight since in practice
\begin{equation*}
|\mathbf{a}^T_i \mathbf{d}_j| << \frac{\mu}{\sigma_\mathrm{y}[i]} \quad \forall \{ (i,j) : i\neq j, i \in K, j \notin K \},
\end{equation*} 
and
\begin{equation*}
|\mathbf{a}^T_i \mathbf{e}| << \varepsilon_{\mathrm{s}} \quad \forall i,
\end{equation*} 
where we denote $\mathbf{a}_i = \mathbf{W}_0 \mathbf{d}_i$.
Good support recovery can be achieved in practical cases with sufficient separation for much higher values of $\mu$, $s$ and $\frac{|\mathbf{x}_i|_\mathrm{min}}{|\mathbf{x}_i|_\mathrm{max}}$, as demonstrated in Section \ref{sec5}.

Stable recovery of Hard Thresholding  in the noiseless case is acquired as long as \cite{Elad}
\begin{equation}\label{3.19}
\min_{i \in K} |\mathbf{d}_i^T\mathbf{y}| > \max_{j \notin K} |\mathbf{d}_j^T\mathbf{y}|.
\end{equation}
Leading to
\begin{equation}\label{3.20}
|\mathbf{x}|_{\mathrm{min}} - (s-1) \mu |\mathbf{x}|_{\mathrm{max}} > \beta_1 >  s \mu |\mathbf{x}|_{\mathrm{max}}.
\end{equation}
In other word,
\begin{equation}\label{3.21}
\frac{|\mathbf{x}|_{\mathrm{min}}}{|\mathbf{x}|_{\mathrm{max}}}> (2s-1)\mu.
\end{equation}
Suppose $s=1$, meaning that there can only be one atom in each stripe, than we have 
\begin{equation}\label{3.22}
\frac{|\mathbf{x}|_{\mathrm{min}}}{|\mathbf{x}|_{\mathrm{max}}} > \mu .
\end{equation}
Practically, in many convolutional models, if the stride is not large enough, this condition is not met, leading us to the following theorem.

\begin{theorem}[Support Recovery using RFN under Sufficient Separation]
\label{Theorem 2}
Perfect recovery in the noiseless case where $s=1$ with RFN is guaranteed for any ratio
\begin{equation*}
0 < \frac{|\mathbf{x}|_{\mathrm{min}}}{|\mathbf{x}|_{\mathrm{max}}} \leq 1 .
\end{equation*}
\end{theorem}
\textit{Proof:} see Appendix B.

It is important to emphasize that in most practical cases, even if the above assumptions do not hold, it does not mean the support detection will fail. It solely means that we cannot guarantee, by means of the following proofs, perfect support detection at the first iteration.

Denote $\mathbf{D}_i \in \mathbb{R}^{L_{\mathrm{h}} \times mL_{\mathrm{s}}}$ as the submatrix partial dictionary yielding point $\mathbf{y}[i]$, that is obtained by restricting the dictionary $\mathbf{D}$ to the support of $m L_\mathrm{s}$ samples equally distributed around point $i$.

\begin{theorem}[Support Recovery with Attenuating RFN kernel in the Noiseless case]
\label{Theorem 3}
Let $\mathbf{y}=\mathbf{Dx}$, where $\mathbf{D}$ is a convolutional dictionary, with normalized atom's variance $\| \mathbf{d}_i \|=\sigma_\mathrm{d} = 1\ \forall i$. Assume a symmetric strictly decreasing receptive field normalization kernel, that is $h[|k|]>h[|k+1|]$ (e.g., a truncated Gaussian window), of length $L_{\mathrm{h}}=L_{\mathrm{d}}$, such that $\mu(\mathbf{HD}_i) << \mu(\mathbf{D}), \ \forall i \in [1, L_{\mathrm{y}}]$. 
Assuming that
\begin{enumerate}
\item K obeys the minimal separation condition with a separation constant $\nu$.
\item
\begin{flalign}\label{23}
\nonumber
\frac{\sqrt{s} \mu}{h_\mathrm{d,min}} < \beta_1 <
1 - \big( h_\mathrm{d}(\nu)  + \mu \big) \frac{\| \mathbf{x}_{-i} \|_1}{|\mathbf{x}[i]| + h_\mathrm{d}(\nu) \| \mathbf{x}_{-i} \|_1}\\
 \quad \forall i \in K,
\end{flalign}
\end{enumerate}
where $h_\mathrm{d}(\nu) \triangleq \max_{p \in [1,m]} \mathbf{H}_\mathrm{d}[\Delta_k+(p-1)L_\mathrm{s},\Delta_k+(p-1)L_\mathrm{s}]$, such that $\mathbf{H}_\mathrm{d}$ is a diagonal matrix holding the attenuated atoms $\ell_2$ norms at the support around some point data at index $i$. Namely $\mathbf{H}_\mathrm{d}[k,k]=\|\mathbf{Hd}^i_k\|_2, \ k \in S^h_i$, where $\mathbf{d}^i_k \in \mathbb{R}^{L_{\mathrm{h}} \times 1}$ denotes the $k$th atom of the submatrix $\mathbf{D}_i$. In other words $h_\mathrm{d}(\nu)$ is the maximal $\ell_2$ norm of an atom's shift at the minimal separation distance multiplied by the RFN window. Recall that we defined $\Delta_k$ as the minimal number of samples between adjacent events. On the other hand, $h_\mathrm{d,min} \triangleq \min_k \mathbf{H}_\mathrm{d}[k,k]$ is the lowest $\ell_2$ norm of a shifted atom multiplied by the RFN window centered around some point $i$, and $\| \mathbf{x}_{-i} \|_1 \triangleq \sum_{t \in S^i_h, t \neq i} |\mathbf{x}[t]| = \|\mathbf{x}_i\|_1 - |\mathbf{x}[i]|$ is the sum of weights affecting the $i$th stripe, besides the spike at location $i$. Then, the support of $\mathbf{x}$ is perfectly recovered with receptive-field normalization thresholding,
\begin{equation*}
\mathrm{Supp}(\mathbf{x}_1)=\mathrm{Supp}(\mathbf{x}),
\end{equation*}
where $\mathrm{Supp}(\cdot)$ is the support of a vector, and $\mathbf{x}_1$ is the recovered sparse vector within one iteration of Algorithms 1-4.
\end{theorem}
\textit{Proof:} see Appendix C.

The suggested approach and the above theorems shed a light on the role of spatial-stride and maxpooling in CNNs \cite{Geron:2017}. When using a spatial-stride, we convolve an input signal is a certain layer, while skipping a fixed number of spatial locations, primarily in an attempt to reduce the computational burden. Consequently, the mutual coherence of the stride convolutional dictionary is smaller, guaranteeing more accuracy and less redundancy at the activation stage. Alternatively, when the atoms shifts are small, usually due to a high sampling rate, the mutual coherence of the convolutional dictionary is relatively high. Naturally, the projection of a shifted atom on the signal may yield strong values in a small area around its true location in the signal. In this case threhsolding can lead to a group of spikes around each true spike. In their work, Gregor and Lecun \cite{Lecun:2010} refer to these issues as ``mutual inhibition and explaining away". One possible way to address this issue in CNNs is by maxpooling. Namely, by taking into account only the maximal or average value in a small local samples neighborhood. In the above method, we propose another possible way to address this issue. That is, one can suppress the strong expressiveness of one component over the other by using an attenuating normalization window, such as a Gaussian window (as opposed to an averaging window), and by fine tuning of the applied thresholds.

The proposed method is closely related to local response normalization (LRN) \cite{Alexnet:2012}. LRN normalizes each neuron's response with respect to the sum of the squared responses of adjacent kernel maps at the same spatial position. The idea, inspired by natural processes in biological neurons, is that strong neuron responses will inhibit weaker neurons responses. In terms of neural nets training, LRN reduces feature maps similarity and encourages competitiveness, which in turn can improve generalization. Practically, the main difference is that we propose to locally normalize the input signal, with respect to spatially close inputs, before applying a filter, in an attempt to scale all responses to the same range, and to prevent stronger spikes from ``casting" a shadow over weaker spikes, both locally and globally, that is, across all of the signal.

\section{Experimental Results}\label{sec5}
In the following two sections, we provide synthetic and real data examples demonstrating the performance of the proposed technique. Hereafter, we refer to the proposed method as RFN-ITA (receptive field normalization iterative thresholding algorithm).
\subsection {Synthetic Data} \label{sec5a}

In order to verify the robustness of the proposed method, we conducted extensive experiments in different scenarios.

First, we generated $J=1000$ random reflectivity independent sequences of length $L_\mathrm{x}=60$. Each reflectivity column is modeled as a zero-mean Bernoulli-Gaussian process \cite{Kormylo:1982}, i.e., reflectors times (depths) are independently distributed with Bernoulli probability $p$, and reflectors amplitudes are normally distributed with mean $\mu$ and variance $\sigma^2$. Mathematically, each reflectivity sequence $x^{(l)}[k]$ is described as
\begin{equation*}
x^{(l)}[k]=r^{(l)}[k]q^{(l)}[k], \quad j=1,...,J,
\end{equation*}
where $r^{(l)}[k] \sim \mathcal{N}(\mu,\sigma_\mathrm{r}^2)$ and $q^{(l)}[k] \sim B(p)$. Clearly, $p$ determines the degree of sparsity. In the following experiments $\sigma_\mathrm{r}=3$, and $0.1 \leq p \leq 0.4$, depending on the chosen minimal separation $\Delta_k$ - the minimal separation between two spikes. Here, $\Delta_k$ does not necessarily obey the minimal separation condition.
  
We assume a source waveform $g(t)$ defined as the real-valued Ricker wavelet
\begin{equation}\label{5.1}
	g(t) = \Big(1-\frac{1}{2}\omega_0^2t^2\Big)\exp\Big(-\frac{1}{4}\omega_0^2t^2\Big),
\end{equation}
where $\omega_0$ is the dominant radial frequency \cite{Wang3}, determining
the width of the pulse. The seismic traces are calculated according to (\ref{2.3}), in a time-invariant noise-free environment, with sampling interval $T_{\mathrm{s}}=4$ ms. Since in this case $\mathbf{G}$ is a signature dictionary, we implemented algorithm \ref{alg3} using a truncated zero-mean Gaussian window of length $L_\mathrm{h}$, with variance $\sigma_\mathrm{h}^2$. The thresholds $\beta_1$ and $\beta_2$ are explicitly determined, and for the rest of the iterations we set $\beta_l=0.5\beta_{l-1}$, where $l$ is the iteration number. The algorithms is stopped when $\|\mathbf{x}_l-\mathbf{x}_{l-1}\|_2<10^{-4}$. We allow up to 4 iterations ($N_{\mathrm{it}}=4$). The step size $\alpha_\mathrm{r}=0.5$ is constant for all experiments.

As a figure of merit, we use the correlation coefficient between the true reflectivity and the recovered reflectivity,
\begin{equation*}
\rho_{\mathbf{X},\hat{\mathbf{X}}} = \frac{\mathbf{X}_{\mathrm{cs}}^T\hat{\mathbf{X}}_{\mathrm{cs}}}{\|\mathbf{X}_{\mathrm{cs}}\|_2 \|\hat{\mathbf{X}}_{\mathrm{cs}}\|_2},
\end{equation*}
where $\mathbf{X}_{\mathrm{cs}}$ and $\hat{\mathbf{X}}_{\mathrm{cs}}$ are column-stack vectors of the true reflectivity image and the recovered reflectivity image, respectively. Table 1 presents the estimated reflectivities score at the first iteration and at the final iteration, and the average number of iterations per trace with respect to the wavelets width, the minimal separation distances and the algorithm parameters $\beta_1, \beta_2, L_\mathrm{h}, \sigma_\mathrm{h}$. Usually $\beta_1$ is close to 1, depending on the minimal separation. Notice that in these experiments the mutual coherence is relatively high:
$ \mu (\mathbf{G}) = 0.764, 0.585$ for $\omega_0=50\pi,80\pi$ respectively.

\begin{figure}[t]
\vspace{-0.5cm}
\centering
\includegraphics[scale=0.55]{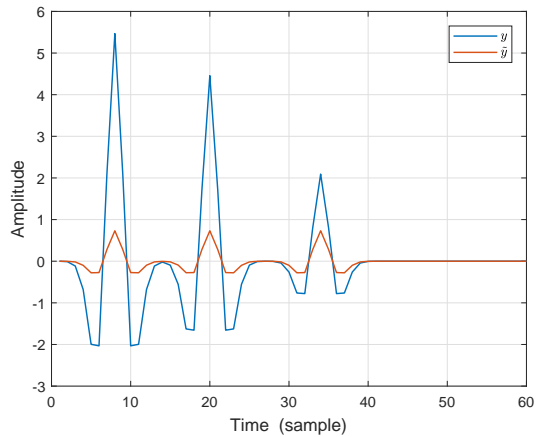} 
\caption{Synthetic data Receptive Field Normalization example: (a) seismic trace; (b) seismic trace after applying receptive-field normalization, with an averaging window of size $L_\mathrm{h}=15$.}
\label{fig1}
\vspace{-0.5cm}
\end{figure}

Figure~\ref{fig1} presents a simple example of a trace after partial receptive-field normalization $\tilde{\mathbf{y}} = \mathbf{W}_0\mathbf{y}$,
with $\omega_0=80\pi$, and $\Delta_k=5$ samples. In this example, the pulses are intentionally completely separated. As can be clearly seen, when the pulses are sufficiently separated, the normalized signal is perfectly balanced regardless of the original local energy. Also, the pulses shape is preserved. Therefore, beyond the theoretical analysis, it can be intuitively comprehended why given sufficient separation, all spikes locations can be detected by simply projecting the dictionary on the normalized signal and applying a threshold, without considering the ratio between the minimal and the maximal coefficients in absolute value in $\mathbf{x}^{(l)}$. Hence, all iterations beyond the first one are required only when reflectors are insufficiently separated with respect to the wavelet's length and the mutual coherence.

\begin{figure}[t]
\centering
\begin{tabular}{c}
\includegraphics[width=6cm, height=2.5cm]{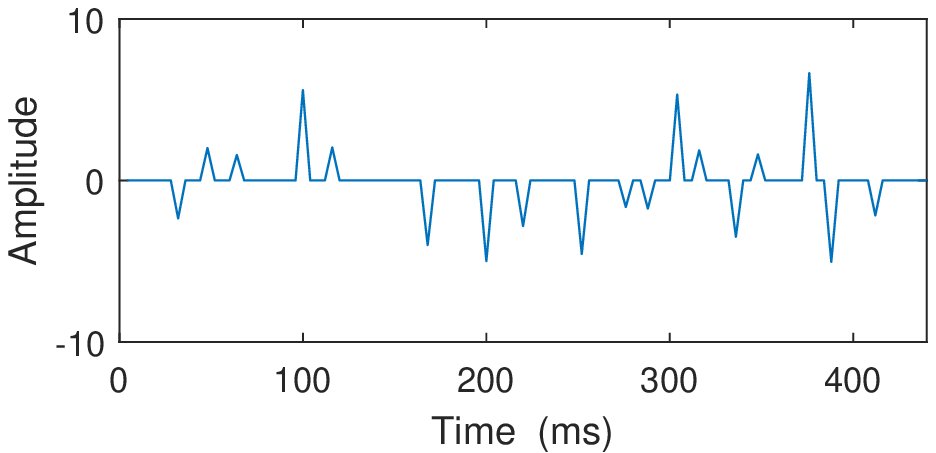} \\
(a) \\
\includegraphics[width=6cm, height=2.5cm]{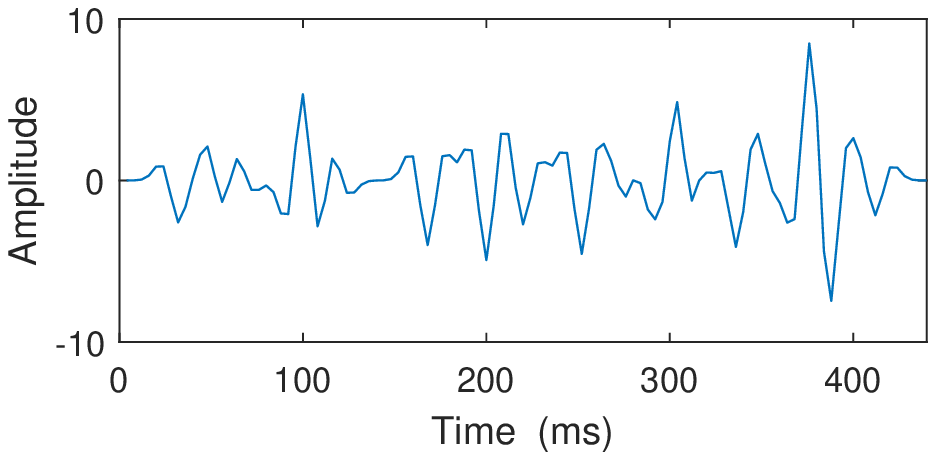} \\
(b)\\
\includegraphics[width=6cm, height=2.5cm]{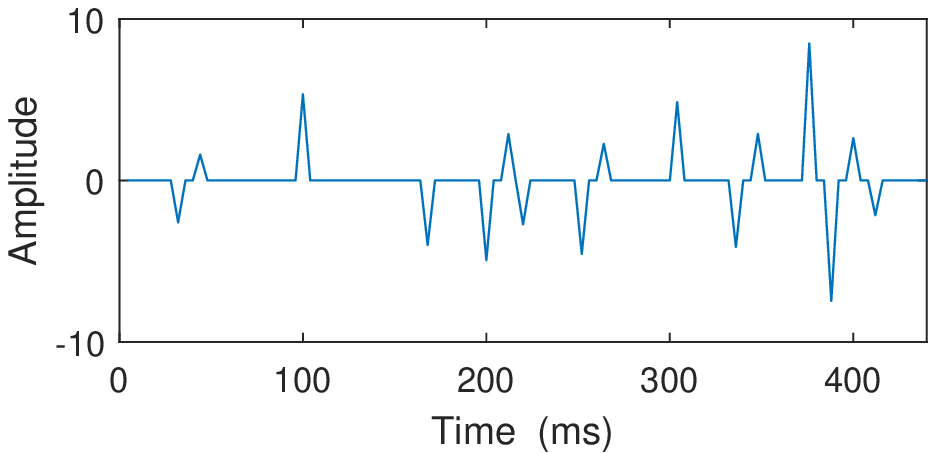} \\
(c)\\
\includegraphics[width=6cm, height=2.5cm]{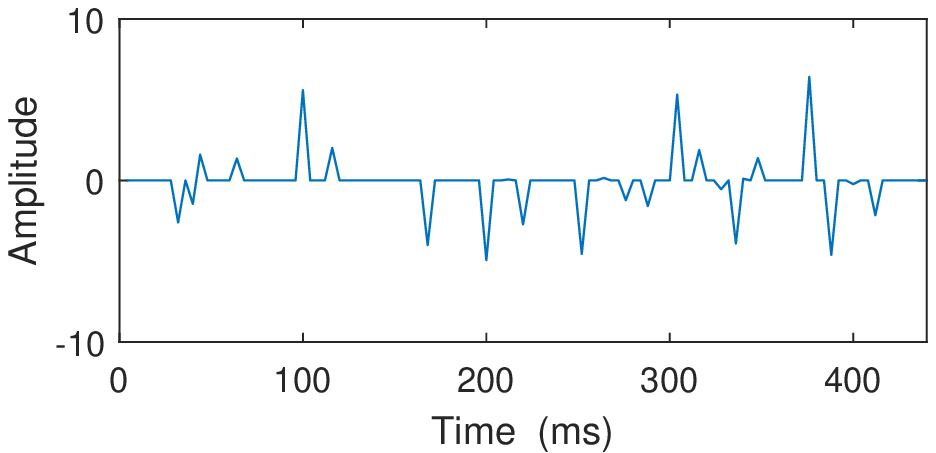} \\
(d) \\
\end{tabular}
\caption{1D synthetic test: (a) True reflectivity; (b) Synthetic trace with $40$~Hz Ricker wavelet; (c) Recovered 1D channel reflectivity signal, first iteration $\rho^1_{\mathbf{X},\hat{\mathbf{X}}}=0.91$;
(d) Recovered 1D channel of reflectivity signal $\rho_{\mathbf{X},\hat{\mathbf{X}}}=0.98$.}
\label{fig2}
\end{figure}

Figure~\ref{fig2}(a) presents an example of a reflectivity channel of length $L_\mathrm{x}=110$ samples ($T_{\mathrm{s}}=4$ms), with $p=0.4$, $\sigma_\mathrm{r}=3$ and a minimal separation distance of $\Delta_k=3$ samples.
The corresponding seismic trace with $\omega_0=80\pi$ is depicted in Fig.~\ref{fig2}(b), and the corresponding estimated reflectivity with $\beta_1=0.95,\beta_l=0.9,2 \leq l \leq 4 $ and step size $\alpha_\mathrm{r}=1$ is presented in Fig.~\ref{fig2}(d). The estimated reflectivity in the first iteration is presented in Fig.~\ref{fig2}(c).

\begin{table*}[ht]
\begin{center}
\caption{Synthetic example - results and parameters: wavelet's scaling $\omega_0$, minimal separation constant $\nu$, thresholds $\beta_1$,$\beta_2$, normalization window size $L_\mathrm{h}$, normalization window std $\sigma_\mathrm{h}$, recovered reflectivity score at the first iteration $\rho^1_{\mathbf{X},\hat{\mathbf{X}}}$, final recovered reflectivity score $\rho_{\mathbf{X},\hat{\mathbf{X}}}$, average number of iterations $M_{\mathrm{it}}$.}
\label{Table 1}
  \begin{tabular}{ |c|c|c|c|c|c|c|c|c|}
    \hline
    $\omega_0$ 	& $\nu$ & $\beta_1$ & $\beta_2$ & $L_\mathrm{h}$ 	& $\sigma_\mathrm{h}$ & $\rho^1_{\mathbf{X},\hat{\mathbf{X}}}$
		
		& $\rho_{\mathbf{X},\hat{\mathbf{X}}}$  & $M_{\mathrm{it}}$ \\ \hline
		
    $80\pi$ (40 Hz) 	& 5 		&  0.95		&  0.88  		& 11  	& 2	 & 0.97  & \textbf{0.995} &  2.58 \\ \hline
    $80\pi$ (40 Hz)  	& 3 		&  0.95  	&  0.87  	  & 11 	  & 2	 & 0.92	 & \textbf{0.97}  &  2.64 \\ \hline
		$80\pi$ (40 Hz)  	& 1 		&  0.8   	&	 0.66   		& 9  	& 2  & 0.81	 & \textbf{0.89}		&	 3.6	\\ \hline
		$50\pi$	(25 Hz)		& 5 		&  0.98   &	 0.98  		& 17 	  & 3  & 0.93  & \textbf{0.985} &  2.19 \\ \hline
		$50\pi$	(25 Hz)		& 3 		&  0.98  	&	 0.87  		& 17	  & 4	 & 0.83  & \textbf{0.9}   &  2.38 \\	\hline 
  \end{tabular} \\
\end{center}
\end{table*}

\begin{figure}[t]
\centering
\includegraphics[scale=0.45]{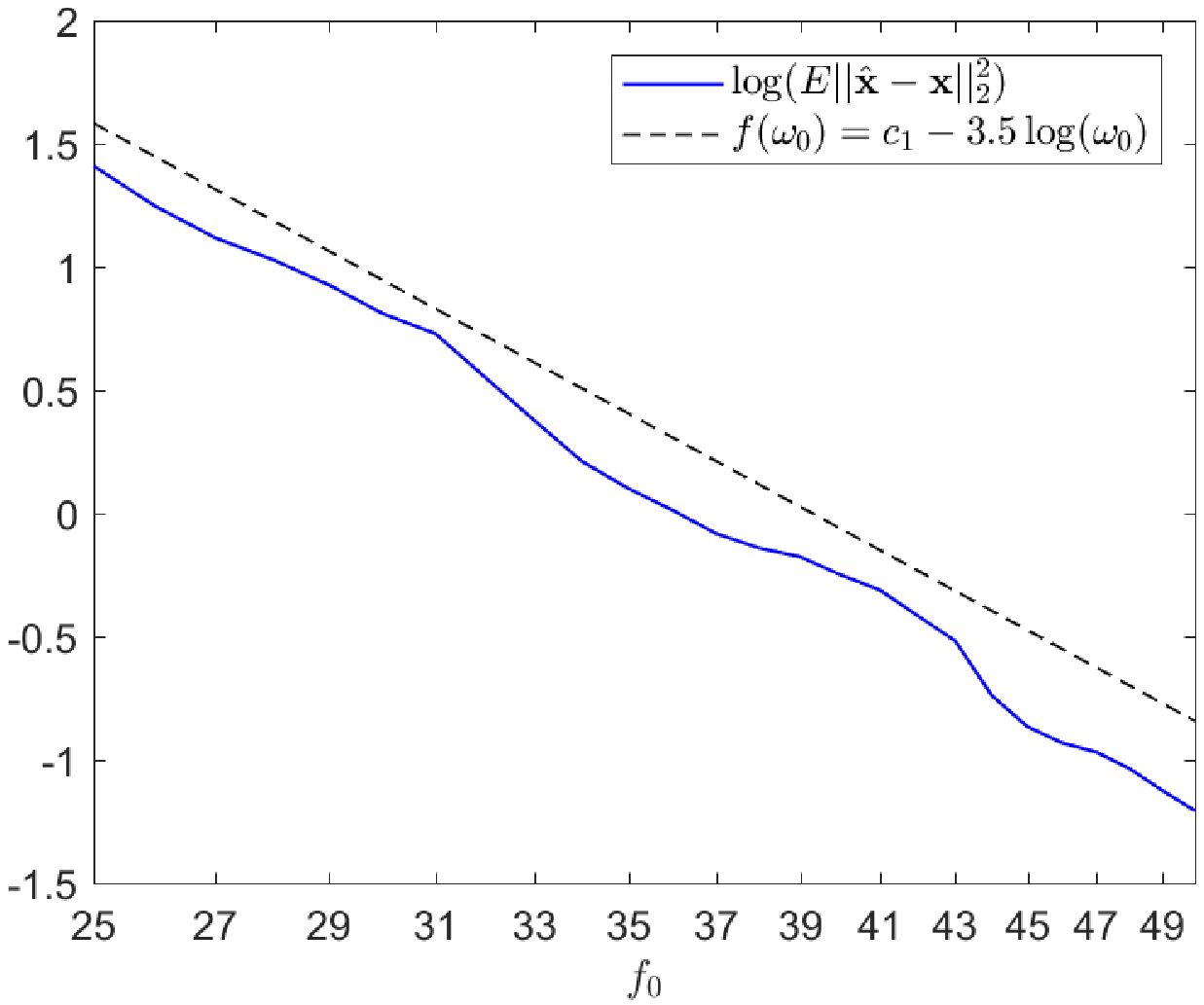} \\
(a) \\

\includegraphics[scale=0.45]{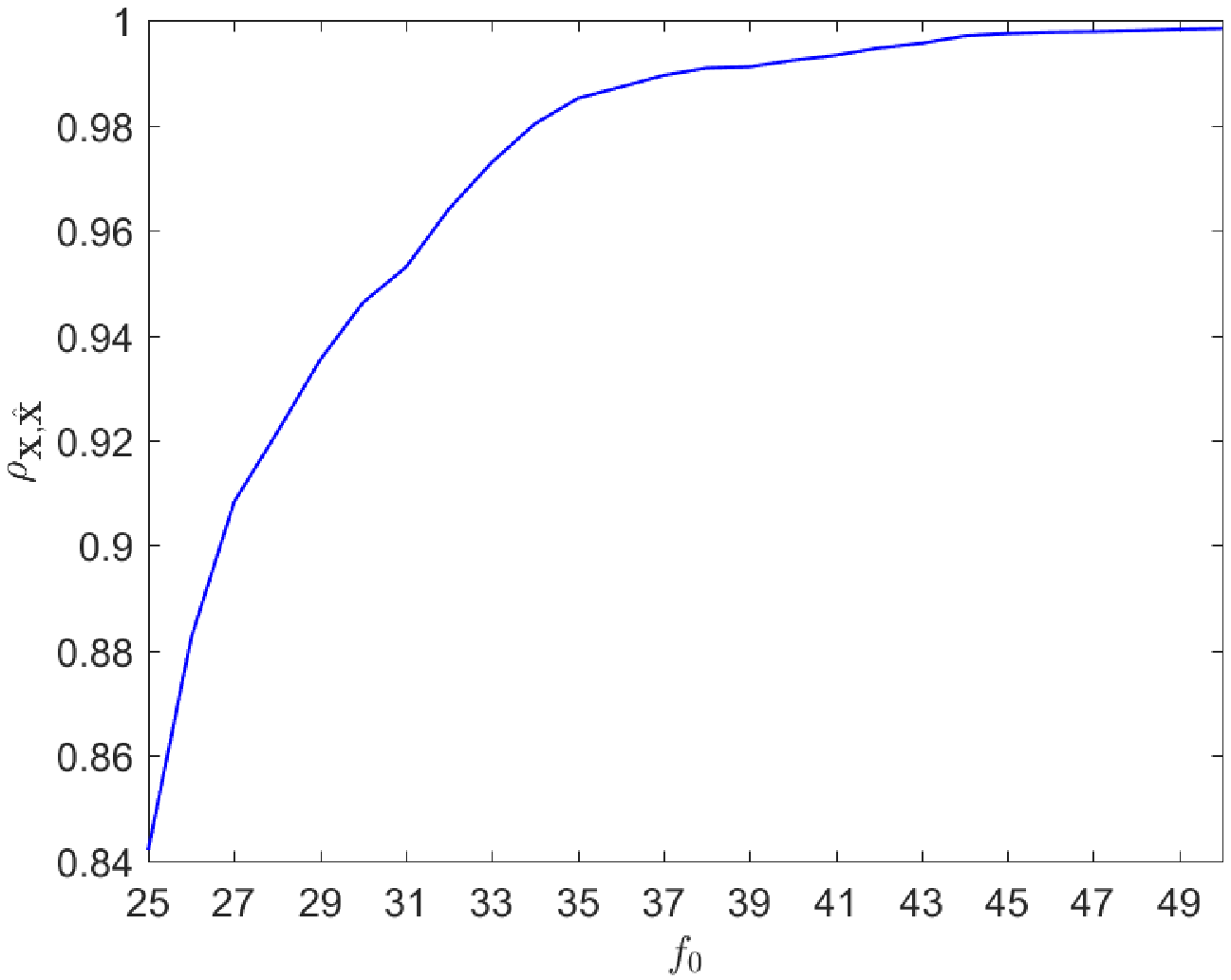} \\
(b)\\
\caption{Reflectivity estimation as a function of the dominant frequency of the Ricker wavelet, SNR=40dB: (a) The blue line corresponds to the log of mean square error $\log (E\|\mathbf{x}-\hat{\mathbf{x}}\|_2^2)$ as a function of $\log(\omega_0)$. The black dashed line corresponds to $f(\omega_0) = c_1 -3.5 \log(\omega_0)$ where $c_1={8F_\mathrm{s}^2\sqrt{L_\mathrm{x}}\sigma_{n}}/{\beta_g}$ is a constant; (b) Correlation coefficient $\rho_{\mathbf{X},\hat{\mathbf{X}}}$ as a function of the wavelet dominant frequency $f_0$(Hz).}
\vspace{-0.5cm}
\label{fig17}
\end{figure}

Figure~\ref{fig17} shows how well one can estimate the reflectivity depending on the dominant frequency of the Ricker wavelet $\omega_0=2\pi f_0$ ranging from 25Hz to 50Hz. Here we used a reflectivity of $J=1200$ channels of length $L_\mathrm{x}=60$ samples ($T_{\mathrm{s}}=4$ms), with $p=0.4$, $\sigma_\mathrm{r}=3$ and a minimal separation distance of $\Delta_k=5$ samples. The seismic data was further destructed by additive noise with $\mathrm{SNR}=40$dB. Here, we used a Gaussian RFN window of length $L_\mathrm{h}=11$ and $\sigma_{\mathrm{h}}=2$. We allow up to 4 iterations where $\beta_1=1.22-0.01(f_0-25),\beta_2=\beta_1+0.2,\beta_l=0.5\beta_{l-1} ,l=3,4$, and $\alpha_\mathrm{r}=0.5$. In Fig. \ref{fig17}(a) the blue line corresponds to the log of the mean square error $\log (E\|\mathbf{x}-\hat{\mathbf{x}}\|_2^2)$ as a function of $\log(\omega_0)$. The black dashed line corresponds to $f(\omega_0) = c_1 -3.5 \log(\omega_0)$ where $c_1={8F_\mathrm{s}^2\sqrt{L_\mathrm{x}}E\|\mathbf{e}\|_2^2}/{\beta_\mathrm{g}}$ is a constant, $\beta_{\mathrm{g}}$ is a parameter that characterizes the concavity of the wavelet as defined in \cite[Definition 2.1]{Pereg:2017A}. As can be observed the mean square error $E\|\mathbf{x}-\hat{\mathbf{x}}\|_2^2$ decreases by a rate of $\omega_0^{3.5}$.
In \cite[Theorem 1]{Pereg:2019A} we prove that under the separation condition the mean square error is inversely proportional to $\omega_0^2$. Hence, the estimation error produced by the proposed algorithm decreases faster than expected, with respect to the wavelet's frequency, even when the separation condition is not satisfied. Figure~\ref{fig17}(b)
depicts the correlation coefficient $\rho_{\mathbf{X},\hat{\mathbf{X}}}$ as a function of the wavelet dominant frequency $f_0$(Hz).

\section{Real Data}\label{sec6}

\begin{figure*}[ht]
\makebox[\textwidth][c]{
\begin{tabular}{ccc}
\includegraphics[scale=0.48]{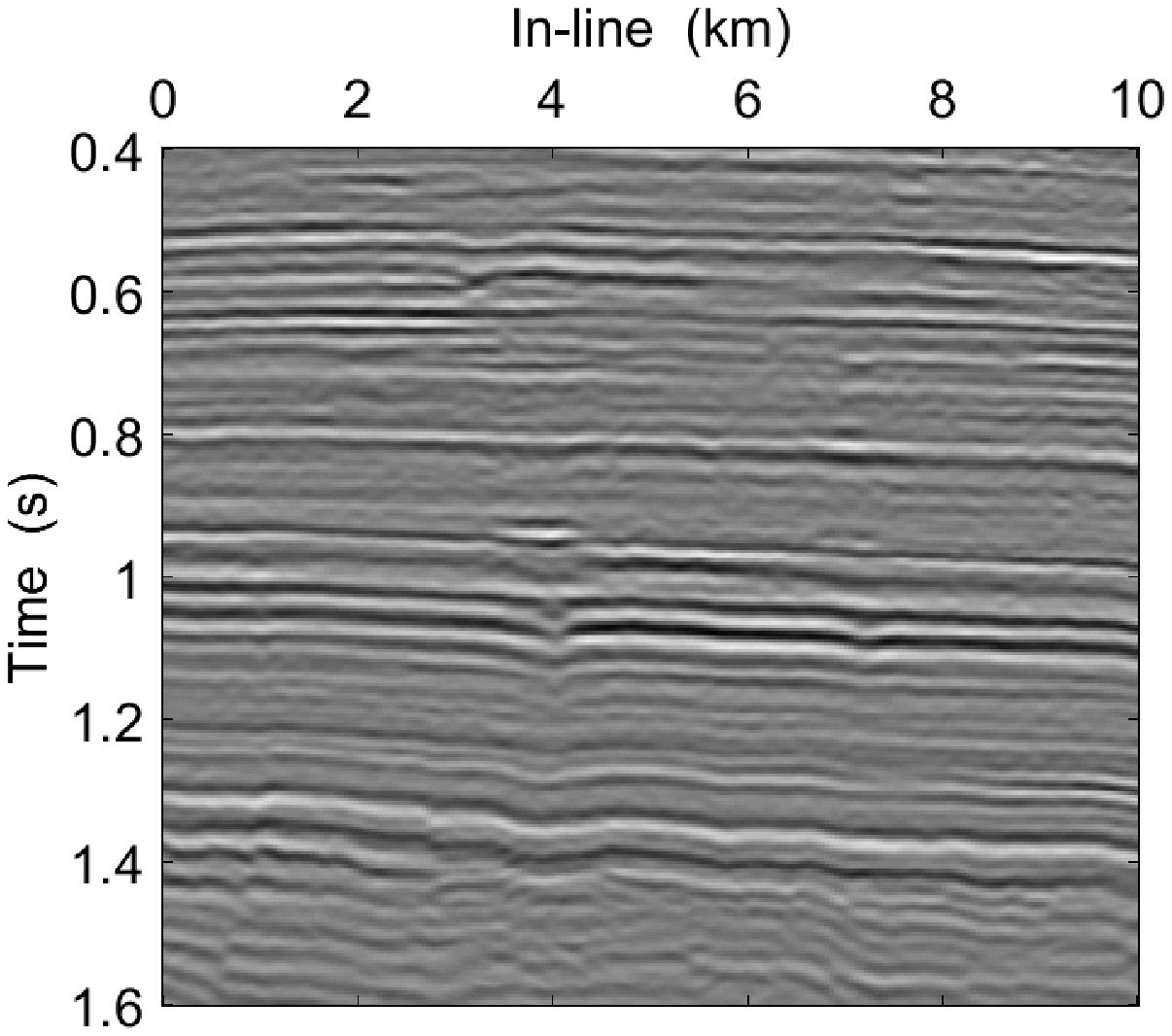}
&
\includegraphics[scale=0.48]{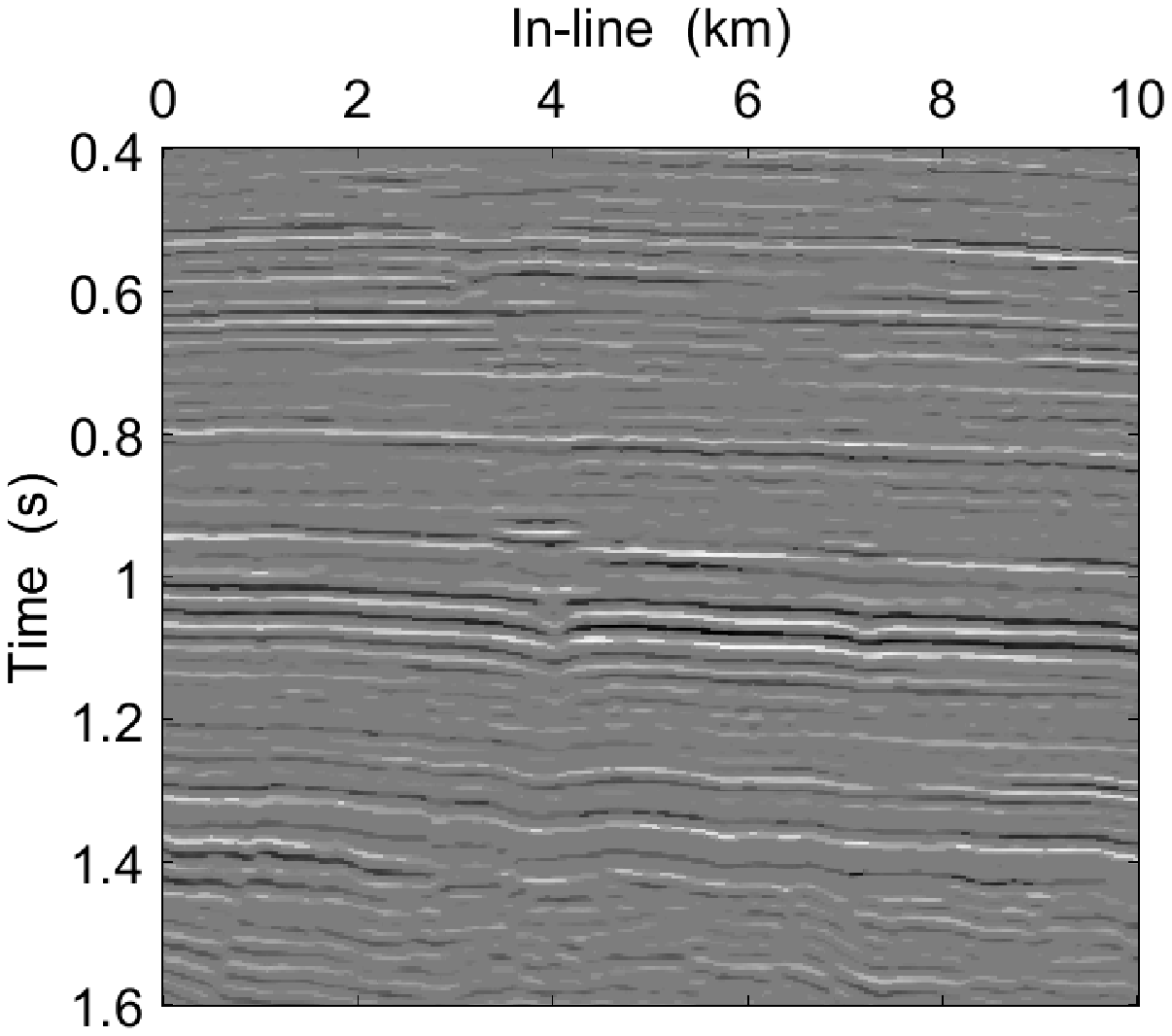}
&
\includegraphics[scale=0.48]{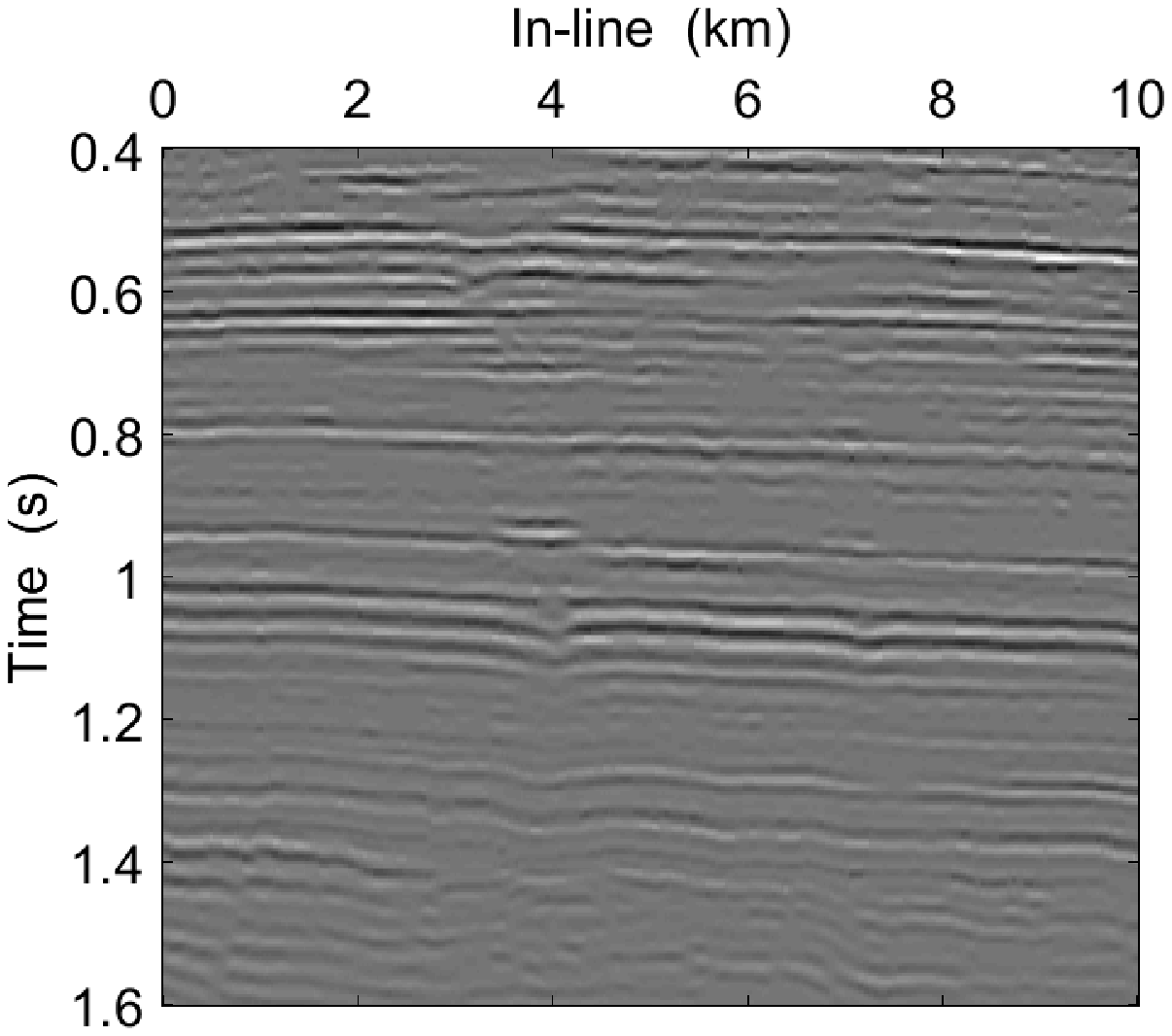}
\end{tabular} }\\
\centering
(a) \hspace{180pt} (b) \hspace{180pt} (c) \\
\caption{Real data inversion results: (a) Seismic data; (b) Estimated reflectivity with $N_{\mathrm{it}}=2$ maximum number of iterations per trace; (c) Reconstructed seismic data, $\rho_{\mathbf{Y},\hat{\mathbf{Y}}}=0.89$.}
\label{fig7}
\end{figure*}

\begin{figure*}[ht]
\makebox[\textwidth][c]{
\begin{tabular}{ccc}
\includegraphics[scale=0.48]{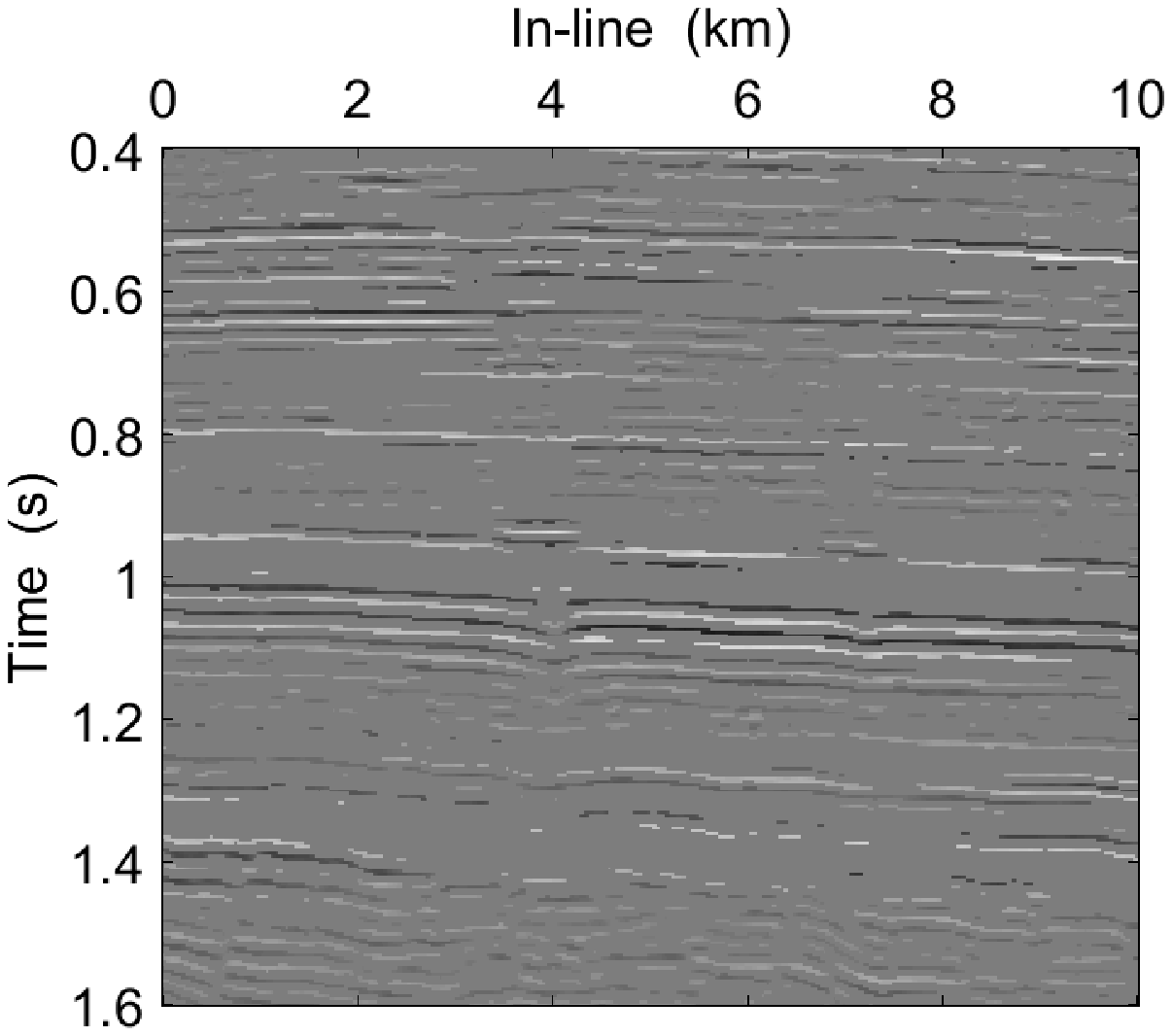}
&
\includegraphics[scale=0.48]{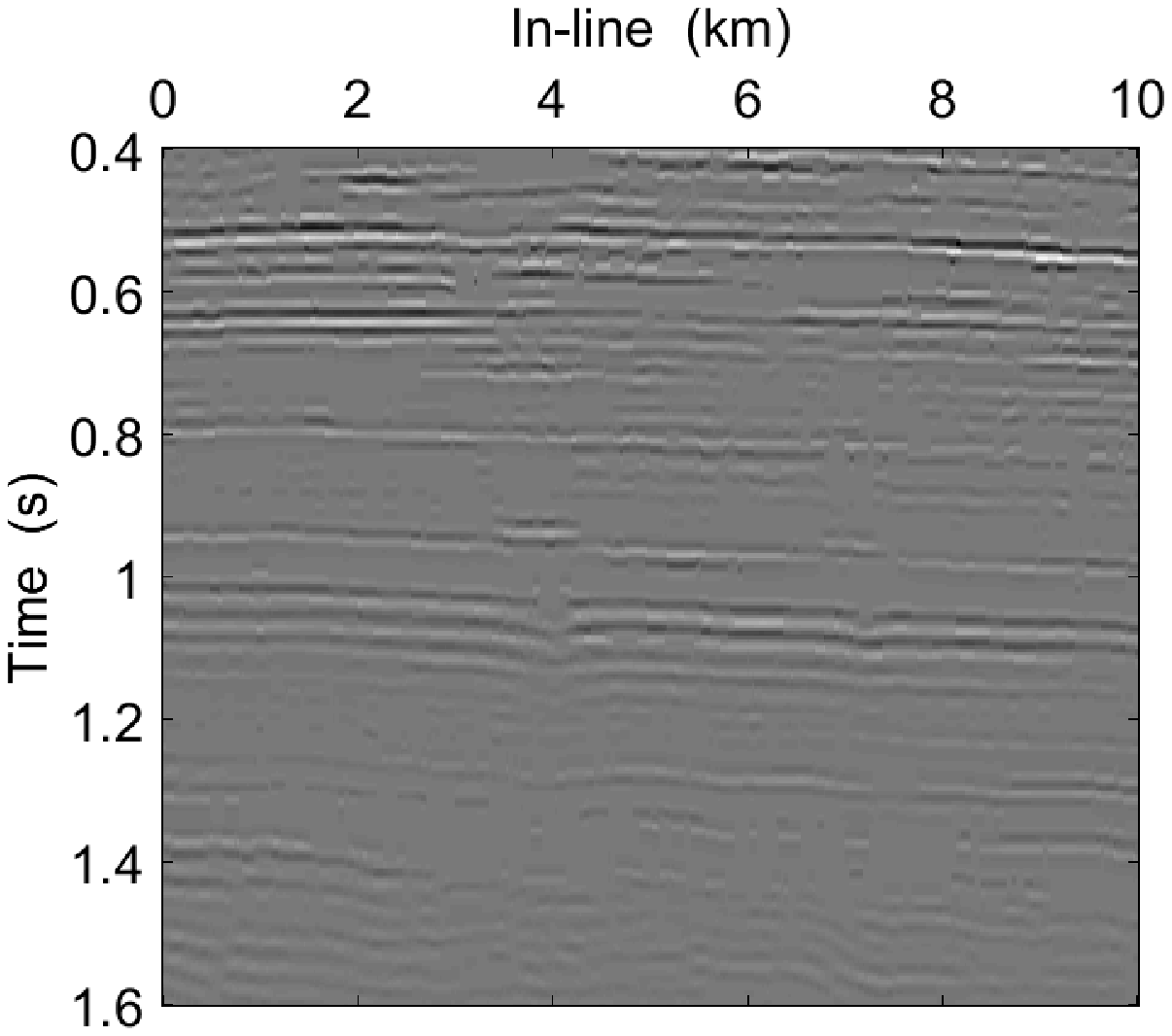}
\end{tabular} }\\
\centering
(a) \hspace{180pt} (b) \\
\caption{Real data inversion results: (a) Estimated reflectivity in the first iteration; (b) Reconstructed seismic data in the first iteration, $\rho^1_{\mathbf{Y},\hat{\mathbf{Y}}}=0.77$.}
\label{fig8}
\end{figure*}

\begin{figure*}[ht]
\makebox[\textwidth][c]{
\begin{tabular}{cc}
\includegraphics[scale=0.48]{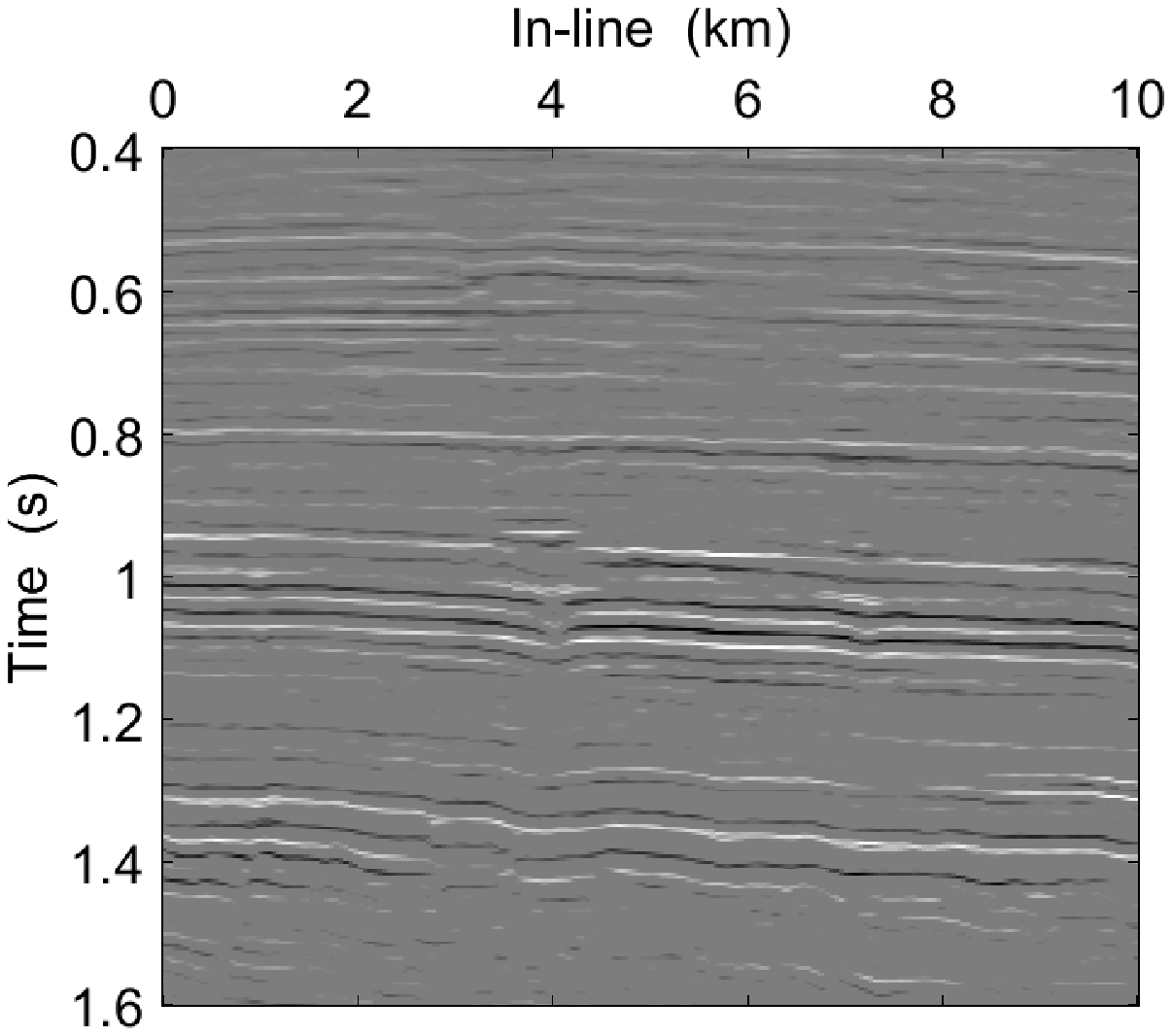}
&
\includegraphics[scale=0.48]{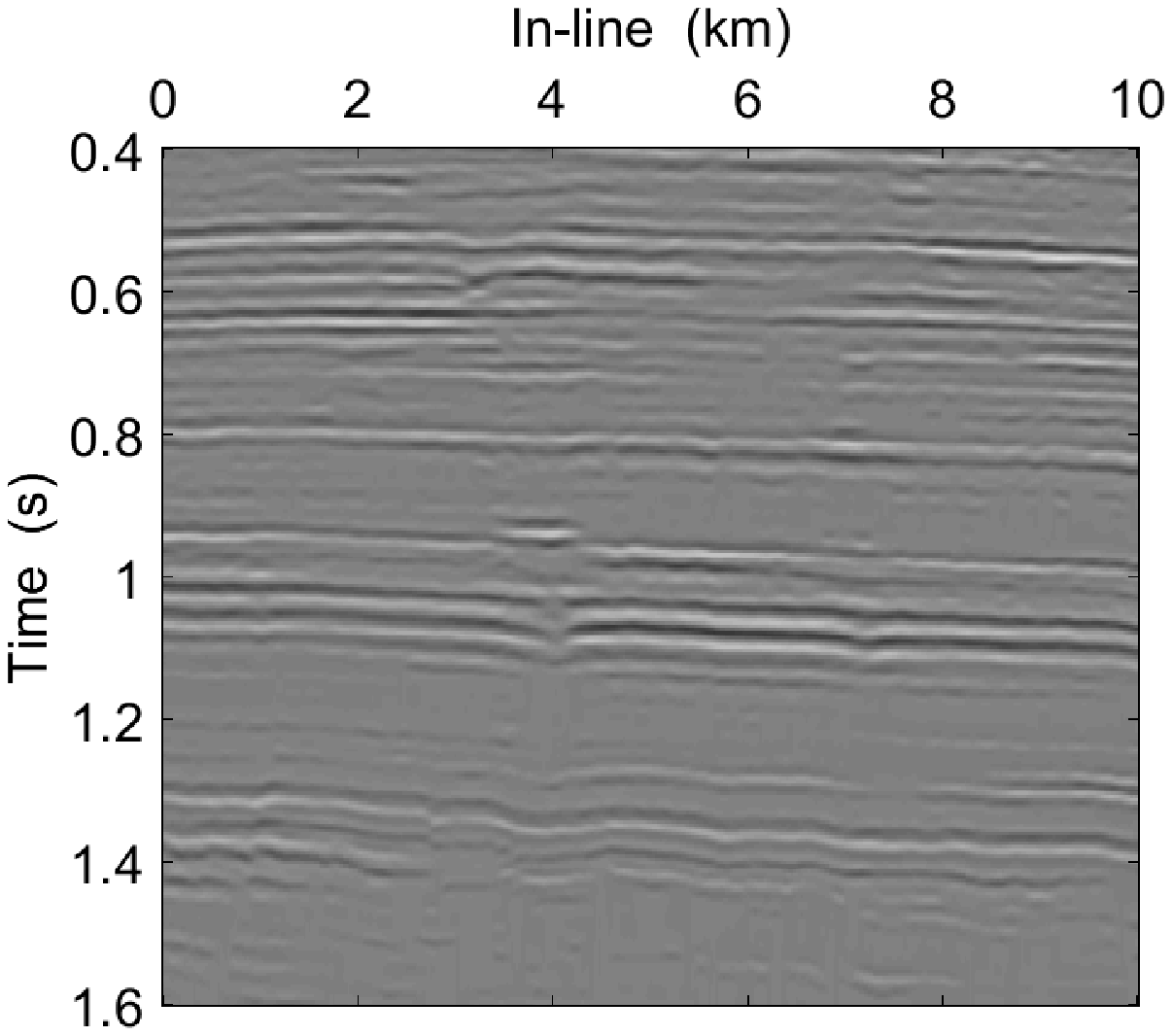}
\end{tabular} } \\
\centering
(a) \hspace{180pt} (b) \\
\caption{Real data inversion results using ISTA: (a) Estimated reflectivity with $M_{\mathrm{it}}=1087$ average number of iterations per trace; (b) Reconstructed seismic data, $\rho_{\mathbf{Y},\hat{\mathbf{Y}}}=0.91$.}
\label{fig9}
\end{figure*}

\begin{figure*}[ht]
\makebox[\textwidth][c]{
\begin{tabular}{cc}
\includegraphics[scale=0.48]{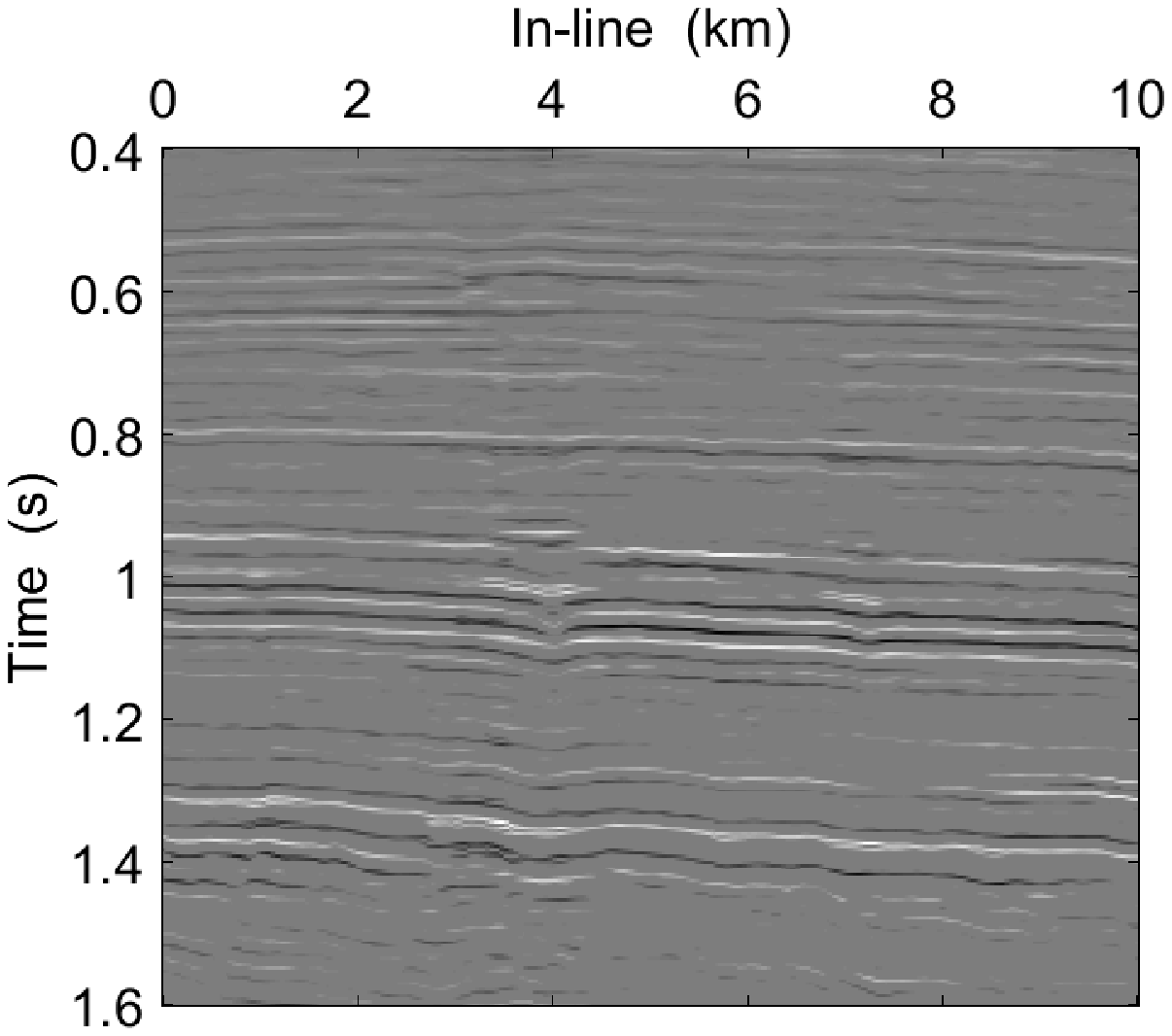}
&
\includegraphics[scale=0.48]{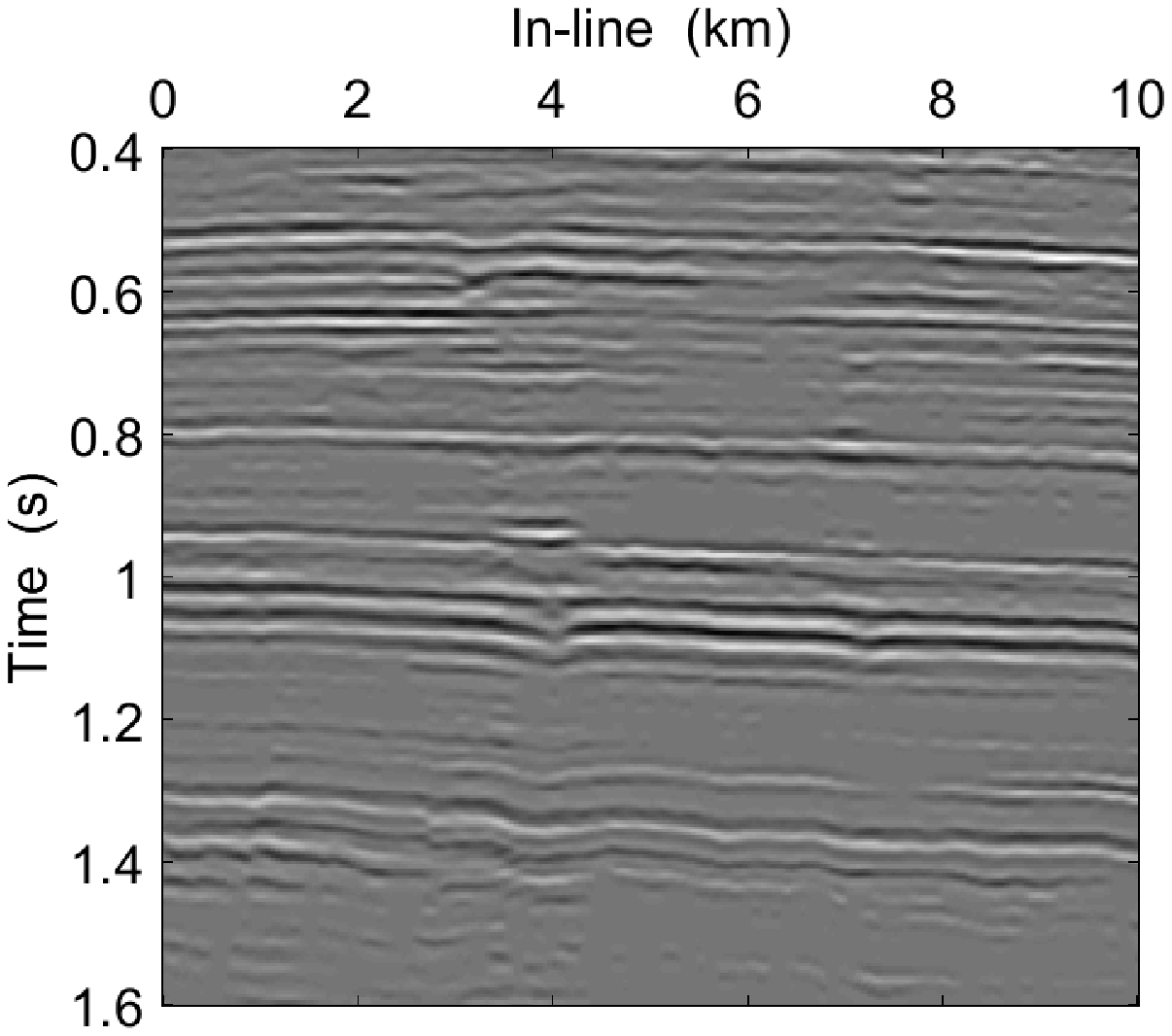}
\end{tabular} } \\
\centering
(a) \hspace{180pt} (b) \\
\caption{Real data inversion results using multichannel time-variant deconvolution \cite{Pereg:2019A} with 3 channels taken into account for each channel estimation, employing LSE discontinuity measure \cite{Cohen:2002}: (a) Estimated reflectivity; (b) Reconstructed seismic data, $\rho_{\mathbf{Y},\hat{\mathbf{Y}}}=0.92$.}
\label{fig18}
\end{figure*}

\begin{figure*}[ht]
\makebox[\textwidth][c]{
\begin{tabular}{ccc}
\includegraphics[scale=0.5]{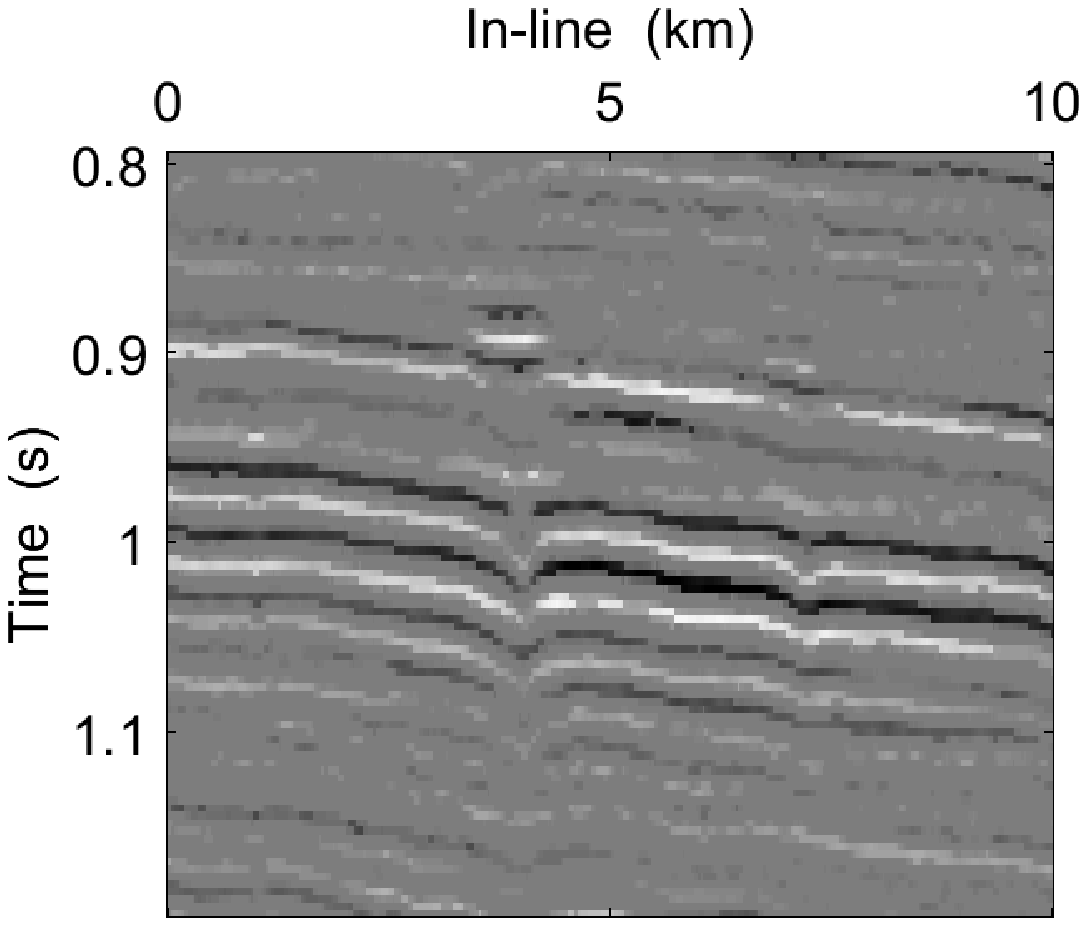}
&
\includegraphics[scale=0.5]{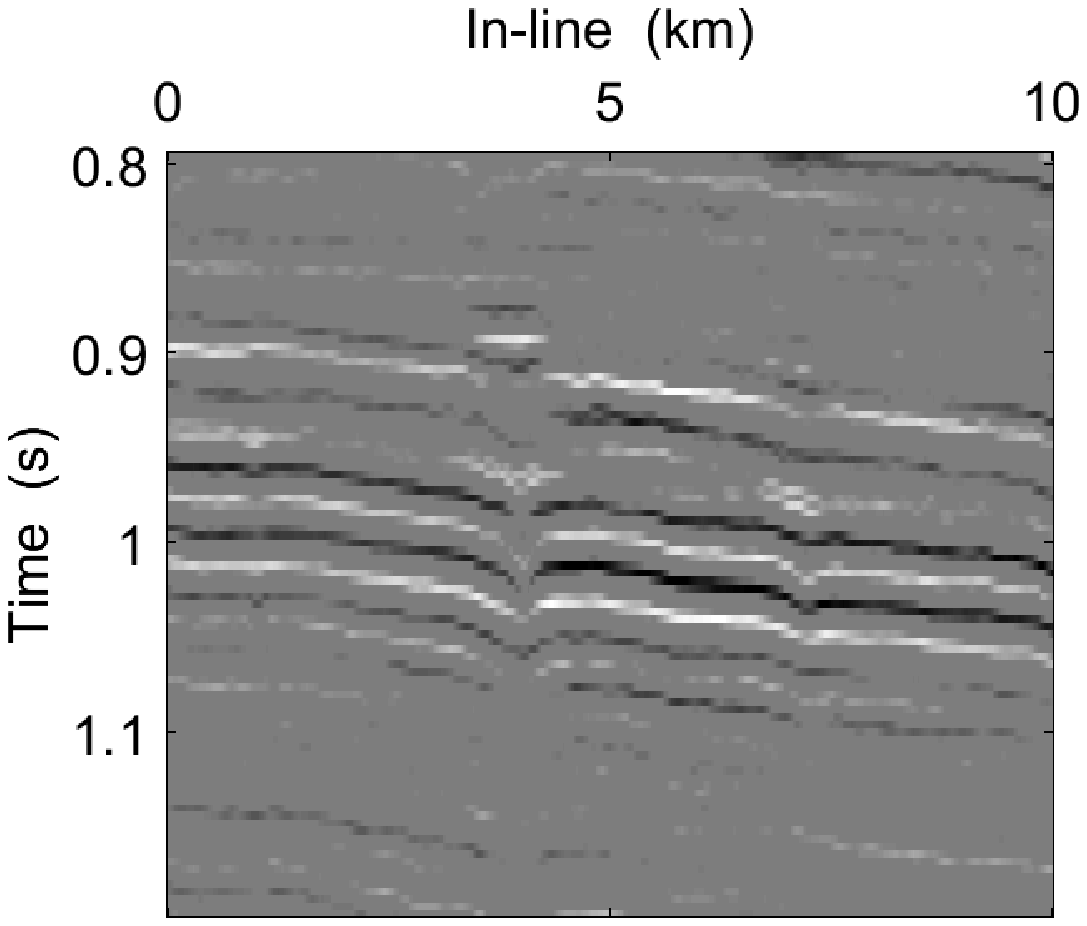}
&
\includegraphics[scale=0.5]{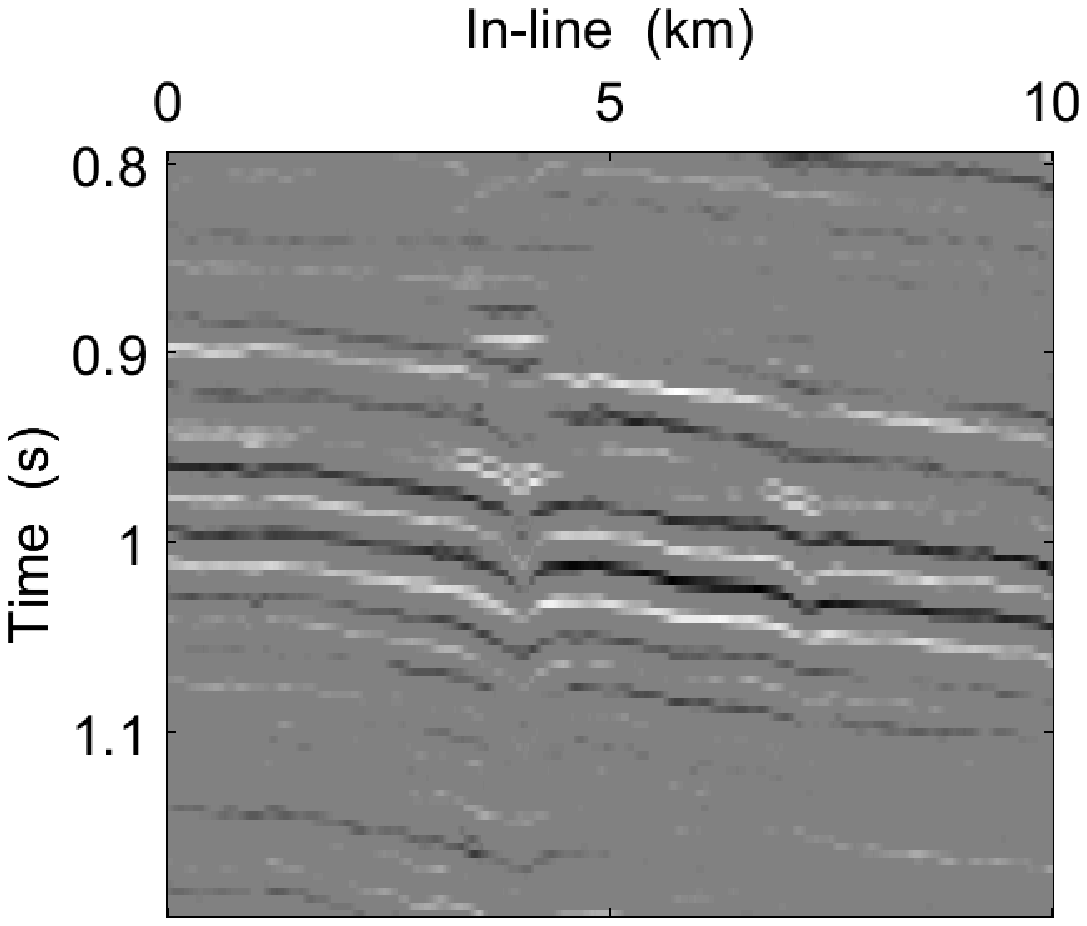} 
\end{tabular}  }\\
\centering
(a) \hspace{160pt} (b) \hspace{160pt} (c) \\
\caption{Zoom into the predicted reflectivities in Figs.~\ref{fig7}-\ref{fig18}: (a) RFN-ITA; (b) ISTA; (c) Multichannel time-variant deconvolution.}
\label{fig10}
\end{figure*}

We applied the proposed method, to real seismic data from a small land 3D survey (courtesy of GeoEnergy Inc., TX). A small 2D seismic image that consists of 400 traces is used for demonstration. Each trace is 1.2 s in duration (300 time samples), with $4$ ms sampling rate. The seismic image is shown in Fig.~\ref{fig7}(a). As can be observed, the data is non-stationary, reflectors are closely spaced and not sufficiently separated and reflectors amplitudes are unbalanced. Assuming an initial Ricker wavelet with $\omega_0=80\pi$ ($40$Hz), we estimated $Q=200$ as described in \cite{Ulrych2}, and derived the set of time-variant pulses $\{g_{\sigma,n}\}$ (see \cite{Pereg:2017A},Appendix D). The estimated reflectivity image and the corresponding reconstructed seismic image using RFN-ITA, at the first iteration and at the second (and final) iteration are depicted in Fig.~\ref{fig7} and Fig.~\ref{fig8}, respectively. For the first iteration we set $\beta_1=1$ and $\tau_1=0.4$. For the second iteration we set $\beta_2=0.7$, $\tau_2=1$ and $\alpha_\mathrm{r}=0.3$. The normalization window is a Gaussian window of length $L_\mathrm{h}=9$ with $\sigma_\mathrm{h}=2$. 

Since the ground truth is unknown, in order to quantify the estimation success, we calculate the reconstructed seismic image from the recovered reflectivity, i.e, $\hat{\mathbf{Y}}=\mathbf{G}\hat{\mathbf{X}}$, depicted in Fig.~\ref{fig7}(c). We assess its correspondence with the given data by the correlation coefficient,
\begin{equation*}
\rho_{\mathbf{Y},\hat{\mathbf{Y}}} = \frac{\mathbf{Y}_{\mathrm{cs}}^T\hat{\mathbf{Y}}_{\mathrm{cs}}}{\|\mathbf{Y}_{\mathrm{cs}}\|_2 \|\hat{\mathbf{Y}}_{\mathrm{cs}}\|_2},
\end{equation*}
where $\mathbf{Y}_{\mathrm{cs}}$ and $\hat{\mathbf{Y}}_{\mathrm{cs}}$ are column-stack vectors of the observed seismic image and the reconstructed seismic image respectively. In terms of the correlation coefficient between the predicted data to the observed data, the first iteration achieved $\rho^1_{\mathbf{Y},\hat{\mathbf{Y}}}=0.77$. While the second iteration achieved $\rho_{\mathbf{Y},\hat{\mathbf{Y}}}=0.89$. Increasing the number of iterations and/or decreasing $\beta_l$ or $\tau_l$ can increase the correlation score, on the expense of the reflectors localization resolution, i.e., these parameters settings can produce smeared results of decreased sparsity, which is usually unwanted. As can be seen, reflectors curves, in the reflectivity estimated in the first iteration, are a bit thick due to high mutual coherence values and insufficient spikes separation. Also, some reflectors are missing. Yet, this image is very close to the final one. Using our method most relevant information is recovered in the first iteration.

\begin{figure}[h]
\makebox[0.5\textwidth][c]{
\includegraphics[scale=0.45]{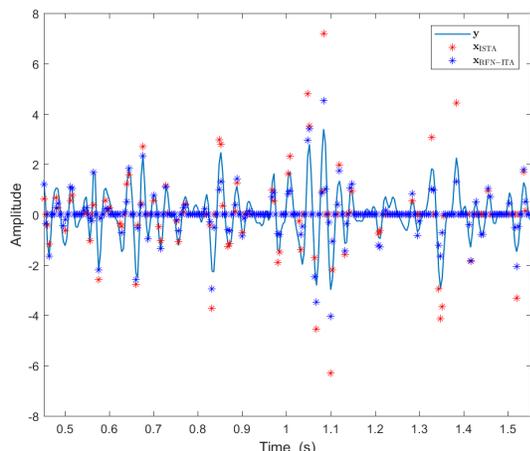} }
\vspace{-1cm}
\caption{Real data results 1D example: seismic trace; ISTA estimated reflectivity, RFN-ITA estimated reflectivity.}
\label{fig16}
\end{figure}

Figures~\ref{fig9}-\ref{fig10} compare the proposed methods results to ISTA results with $\beta=0.14$ and a score of $\rho_{\mathbf{Y},\hat{\mathbf{Y}}}=0.91$, and to the multichannel time-variant method in \cite{Pereg:2019A} with a score of $\rho_{\mathbf{Y},\hat{\mathbf{Y}}}=0.92$. As can be seen, a slightly better score does not necessarily indicate a visually enhanced reflectivity. As can be observed in the predicted reflectivity produced by ISTA, some layers are incomplete, and it appears that some spikes are annihilated. Here, ISTA required an average of 1087 iterations per trace. In this example, RFN-ITA reduces the number of required iterations by a factor of 500. Figure~\ref{fig16} presents one seismic trace at in-line offset 7.5 km, and the corresponding estimated 1D reflectivity by ISTA and by RFN-ITA. As can be visually observed, the data is of inherent ambiguity, and the estimated results differ mostly in amplitude, and share relatively close supports.

Throughout our experiments with ISTA, we observed that decreasing $\beta=\frac{c}{\lambda}$, which decreases the sparsity of the solution, comes at the expense of significantly increasing the number of iterations. Also, as can be seen in the presented examples, ISTA performance is terms of the visual quality of the image, deteriorates with a time-variant dictionary, taking a Q factor into account. Moreover, in practice, ISTA number of iterations increases dramatically with the length of the signal. Another drawback is that the constant $\frac{c}{\lambda}$ depends on the maximum eigenvalue of $\mathbf{D}^T\mathbf{D}$ that is harder to compute at larger scales \cite{Beck:2009}. RFN does not suffer from this complications because an approximate solution is reached within a limited number of iterations, where each data stripe is analyzed locally, yet the signal is updated globally without significant increase in computational complexity.

It is important to emphasize that most real data super resolution problems, and specifically seismic inversion, do not comply with the theoretical bounds constraints (e.g. \cite{Elad,Papyan:2017}) for separation, amplitude balance and noise. Therefore, applying classical algorithms to real data applications is not expected to result in perfect recovery. In \cite{Pereg:2017A} and \cite{Pereg:2019A} we show that under the separation condition in a noise-free environment we can perfectly recover the reflectivity in an attenuating environment, providing that we can estimate $Q$. However, when working with real data there is no guarantee that the true reflectors are sufficiently separated nor that Q is constant or estimated correctly. Also, we observe that the resolution of the estimation results is highly sensitive to user-dependent parameters. Overall, real data is of inherent uncertainty.

Since the proposed method requires only 2-4 iterations per trace, its implementation entails significantly low computational complexity. Solving using MATLAB, the processing time of the above data set of $300\times 401$ on a standard CPU Intel(R)Core(TM)i7-7820HQ@2.90GHz is 122 ms, that is, about 305 $\mathrm{\mu}$s per trace. Whereas the processing time of the same data set for ISTA is 13.13 seconds, that is, 32.8 ms per trace. For the 3D time-variant multichannel method the processing time is 16.06 minutes, and 2.41 seconds per trace. Which means, the proposed method is about 100 times faster than ISTA. This in turn assures us of a promising potential of the suggested method to be incorporated in the future in large scale real-time data processing applications.

\section{Learned RFN-ITA}\label{sec7}
At this stage, a learned version of the proposed method, inspired by LISTA, seems almost inevitable.
Namely, let us unfold $\Theta$ iterations of RFN-ITA, and set the filters building each convolutional dictionary at each iteration $\mathbf{D}_\theta$ as trainable parameters.
In other words, we design a neural net, such that the output at layer $\theta$ is
\begin{flalign}
\nonumber
& \mathbf{\Delta r}_{\theta+1} = \mathbf{y} - \mathbf{D}_{\theta} \mathbf{x}_\theta
\\
\nonumber
& \mathbf{\Delta q}_{\theta+1} = \mathcal{S}_{\beta_{\theta+1}}\Big(\mathbf{D}_{\theta+1}^T \tilde{\mathbf{W}}_{\theta} \mathbf{\Delta r}_{\theta+1} \Big) 
\\
& \mathbf{x}_{\theta+1} = \alpha_{\theta+1} \Big(\mathbf{\Delta q}_{\theta+1} \odot \mathbf{\Delta r}_{\theta+1} \Big) + \mathbf{x}_{\theta}, \quad \theta=0,1,..., \Theta-1,
\end{flalign}
where $\tilde{\mathbf{W}}_{\theta}$ is a scaled RFN matrix as defined in (\ref{3.10}), without dividing by $\|\mathbf{d}_i\|_2$, since at the training stage the dictionary is unknown. In other words, 
\begin{equation}\label{7.1}
\tilde{\mathbf{W}}_{\theta}=\mathrm{diag} \Bigg( \frac{1}{\tilde{\sigma}^{\theta+1}_{\Delta r}[k]} \Bigg), k=1,...,L_{\mathrm{y}}.
\end{equation}
$\Psi=\{\alpha_\theta, \beta_\theta, \mathbf{D}_\theta\}$ are the learned parameters. 
Here we have used the soft-thresholding operation because its implementation using the ReLU function facilitates convergence.
Training is performed using stochastic gradient descent to minimize the loss $\mathcal{L}(\mathbf{y};\Psi)$ between the model predicted code $\mathbf{x}_{\Theta}=f(\mathbf{y};\Psi)$ to a known code over a training set of known $\{\mathbf{y}\}_{p=1}^P$ and the corresponding $\{ \mathbf{x} \}_{p=1}^P$, 
\begin{equation}
\min_\Psi \frac{1}{P} \sum_{p=1}^P \| f(\mathbf{y}_p;\Psi) - \mathbf{x}_p \|^2_2.
\end{equation}
Once trained, the model is expected to produce sparse codes for signals from the same probability distribution, without requiring their original true dictionary.

\begin{figure*}[t]
\vspace{-0.2cm}
\makebox[\textwidth][c]{
\includegraphics[scale=0.7]{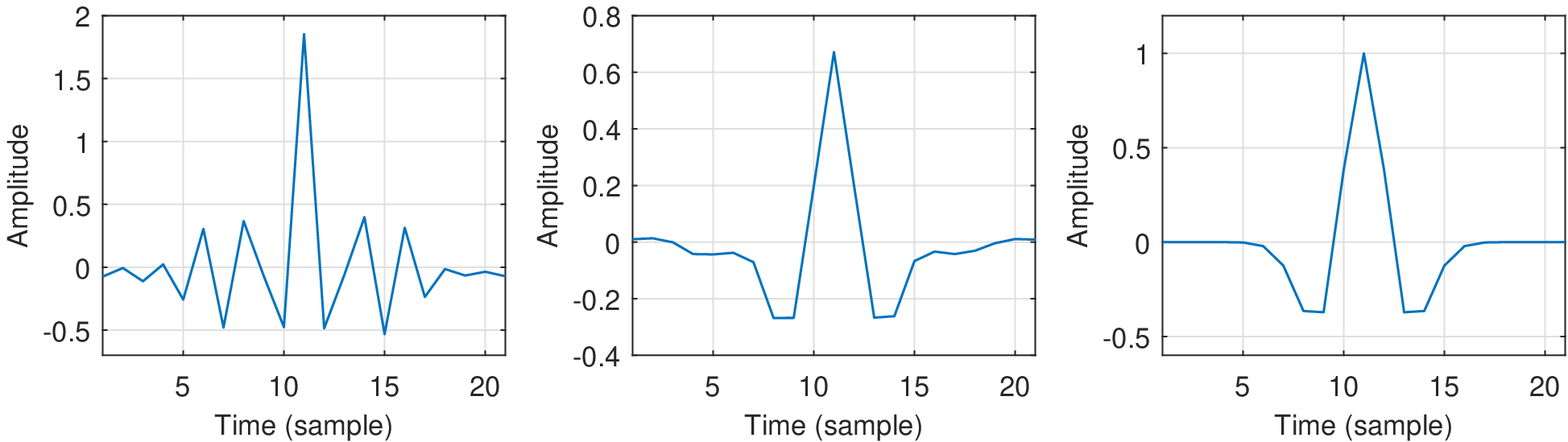}
}\\
\centering
(a) \hspace{120pt} (b) \hspace{120pt} (c)\\
\caption{Learned RFN-ITA parameters: (a) $\mathbf{D}_1$ kernel; (b) $\mathbf{D}_2$ kernel; (c) $80\pi$ Ricker wavelet ($F_{\mathrm{s}}=250$Hz).}
\label{fig13}
\end{figure*}

Figure~\ref{fig13} compares the learned filters to the true wavelet $g[k]$ that was used to create the observed traces, for a two layer model, that is $\Theta=2$, trained to recover only the support, as described in Algorithm \ref{alg4}. In this section we used Pytorch \cite{Pytorch:2019} for the numerical implementation. A training set of $N=997$ random reflectivity independent sequences of length $L_\mathrm{x}=60$, as described in Section~\ref{sec5a}, with $\nu=3$, $\omega_0=80\pi$, and $F_{\mathrm{s}}=250$Hz. Table 2 presents the testing correlation score for this example $\rho^A_{\mathbf{X},\hat{\mathbf{X}}}$, the support correlation score $\rho^{S}_{\mathbf{X},\hat{\mathbf{X}}}$, and the learned parameters, $\beta_1$, $\beta_2$ for a test set of $N=997$ traces, with different minimal separation constants $\nu=5,3,1$. 
We do not force any constraints on the weights parameters. 

\begin{figure}[t]
\vspace{-0.5cm}
\makebox[0.5\textwidth][c]{
\includegraphics[scale=0.55]{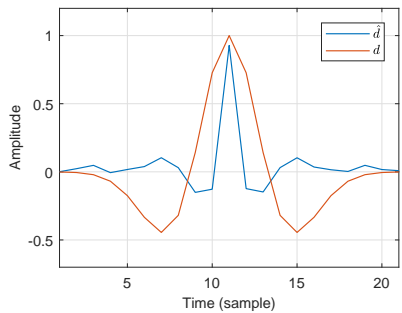}
}
\vspace{-0.5cm}
\caption{Learned RFN-ITA parameters: $\mathbf{D}_1 = \mathbf{D}_2$ compared to original $50\pi$ Ricker wavelet ($F_{\mathrm{s}}=250$Hz).}
\label{fig14}
\vspace{-0.25cm}
\end{figure}

\begin{figure*}[t]
\makebox[\textwidth][c]{
\includegraphics[scale=0.7]{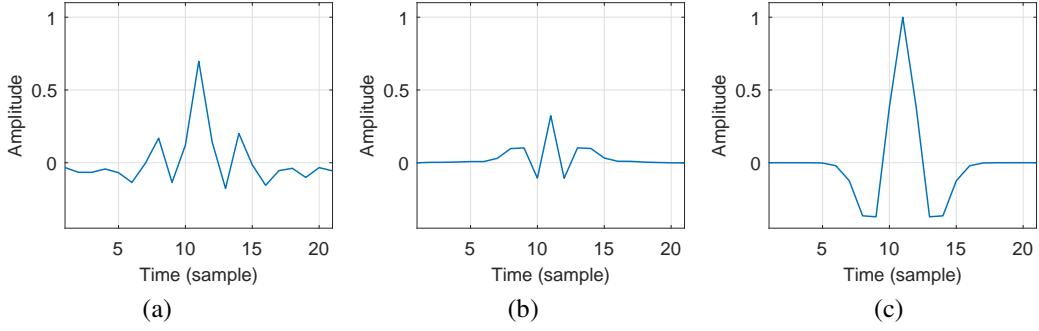}
}\\
\centering
(a) \hspace{120pt} (b) \hspace{120pt} (c)\\
\caption{Learned RFN-ITA parameters: (a) $\mathbf{D}_1$ kernel; (b) $\mathbf{D}_2$ kernel; (c) $80\pi$ Ricker wavelet ($F_{\mathrm{s}}=250$Hz).}
\label{fig15}
\end{figure*}

Figure~\ref{fig15} shows the learned filters versus the true wavelet, for a two layer model, that is $\Theta=2$, trained to recover both the support and amplitude with $\nu=1$, as described in Algorithm \ref{alg3}. Figure~\ref{fig14} presents the learned filter for Algorithm \ref{alg3} when $\nu=1$, $\omega_0=50\pi$ and we force $\mathbf{D}_1=\mathbf{D}_2$. As can be seen, the learned pulse is significantly narrower then the original one, and its side-lobes are shallower. The corresponding correlation scores $\rho^B_{\mathbf{X},\hat{\mathbf{X}}}$ and $\rho^C_{\mathbf{X},\hat{\mathbf{X}}}$, respectively, are also presented in Table 2. As evident, there is no significant variations between the models' performance, since the prediction precision is inherently dependent on sufficient separation between spikes, and the signal bandwidth, represented by $\nu$ and $\omega_0$.

At this stage, we shall leave further work on the learned extension of RFN-ITA, the relevant mathematical analysis and its applications to our future research.

\begin{table*}[t]
\begin{center}
\caption{Synthetic example - results and parameters: wavelet's scaling $\omega_0$, minimal separation constant $\nu$, thresholds $\beta_1$,$\beta_2$, normalization window size $L_\mathrm{h}$, normalization window std $\sigma_\mathrm{h}$, recovered support score $\rho^{S}_{\mathbf{X},\hat{\mathbf{X}}}$, final recovered reflectivity score $\rho^A_{\mathbf{X},\hat{\mathbf{X}}}$ for learned Algorithm \ref{alg4} implementation, final recovered reflectivity score $\rho^B_{\mathbf{X},\hat{\mathbf{X}}}$ for learned Algorithm \ref{alg3} implementation, final recovered reflectivity score for learned Algorithm \ref{alg3} implementation $\rho^C_{\mathbf{X},\hat{\mathbf{X}}}$ with $\mathbf{D}_1=\mathbf{D}_2$.}
\label{Table 2}
  \begin{tabular}{ |c|c|c|c|c|c|c|c|c|c|c|}
    \hline
    $\omega_0$ 	& $\nu$ & $\beta_1$ & $\beta_2$ & $L_\mathrm{h}$ 	& $\sigma_h$ & $\rho^{S}_{\mathbf{X},\hat{\mathbf{X}}}$ &  $\rho^A_{\mathbf{X},\hat{\mathbf{X}}}$ & $\rho^B_{\mathbf{X},\hat{\mathbf{X}}}$ & $\rho^C_{\mathbf{X},\hat{\mathbf{X}}}$ \\ \hline
		
    $80\pi$ (40 Hz) 	& 5 		&  0.90		&  0.57  	& 11  	& 2	 & 0.99  &  0.998 & 0.985 & 0.986	\\ \hline
    $80\pi$ (40 Hz)  	& 3 		&  1.24  	&  0.43  	& 11 	  & 2	 & 0.95	 &  0.97  & 0.96  & 0.96	\\ \hline
		$80\pi$ (40 Hz)  	& 1 		&  0.81  	&	 0.29   & 9  	  & 2  & 0.89	 & 	0.91	& 0.89  & 0.91 	\\ \hline
		$50\pi$ (25 Hz)  	& 5 		&  2.17  	&  0.80  	& 17 	  & 3	 & 0.99	 &  0.99  & 0.985 & 0.987 \\ \hline
		$50\pi$ (25 Hz)  	& 3 		&  2.58  	&  0.43  	& 17 	  & 2	 & 0.94	 &  0.93  & 0.87  & 0.957 \\ \hline
  \end{tabular} \\
\end{center}
\end{table*}

\section{Conclusions}\label{sec8}

We have presented an efficient modification of the classic iterative thresholding algorithms for convolutional sparse coding. We have shown that receptive field normalization leads to a substantial reduction in the number of required iterations to reach an approximate sparse code vector. We demonstrated that whether the dictionary is known or learned via neural network training, about 2-4 iterations are sufficient to produce a sparse code close to the true one, even when the dictionary's mutual coherence is relatively high. The proposed algorithms entail a significantly low computational complexity, and therefore can potentially contribute to a dramatic speed-up in many real-time applications, such as super resolution and pattern recognition systems, as well as neural nets feature extraction in other existing systems. Hopefully, future work will further investigate the proposed methods and their extension to other applications. An extension to seismic blind deconvolution is also possible. The theoretical analysis could be extended to a statistical point of view rather than a worst case scenario approach.

\section*{Appendix A}
\textbf{Proof of Theorem 1}

We introduce the following assumptions:

\textbf{A 1} (Stationary Receptive-Field Energy)
We assume that for every support index $i$ the local weighted energy as defined in (\ref{3.1}) is approximately constant and equal to the energy at $i$, that is 
\begin{equation} \label{A1}
\sigma_\mathrm{y}[k] = \sigma_\mathrm{y}[i] \quad \forall k \in K_i, \ i \in K,
\end{equation}
where $K_i$ is the neighborhood of $i$, and $k$ belongs to the neighborhood $K_i$ provided that $|k-i| \leq \frac{L_\mathrm{d}}{2}-1$.
Assume without loss of generality the dictionary columns are normalized, and
denote $\mathbf{a}_i = \mathbf{W}_0 \mathbf{d}_i$. According to the above assumption we have
\begin{equation}\label{A2}
|\mathbf{a}^T_i \mathbf{d}_i| = \frac{1}{\sigma_\mathrm{y}[i]}  \quad \forall i \in K.
\end{equation}
In other words the corresponding atom after RFN is approximately equivalent to a scaled version of the original atom. Denote $\mathbf{v}_i$ such that $\mathbf{a}_i=\mathrm{diag}^{-1}(\sigma_\mathrm{y}[i])\odot (\mathbf{d}_i-\mathbf{v}_i)$, then (\ref{A2}) is true when either $\mathbf{v}_i^T\mathbf{d}_i=0$ or $\mathbf{v}_i=\mathbf{0}$.
Clearly, assumption \textbf{A 1} is not completely accurate, since for each data sample, the normalization window changes its location, therefore the local energy changes.
However practically we observe that the difference $\| \sigma_\mathrm{y}[i] \mathbf{a}_i -  \mathbf{d}_i \|_2$ is relatively very small and that $\mathbf{v}^T_i\mathbf{d}_i$ can be neglected for the purposes of the following mathematical analysis and for the sake of brevity.

\textbf{A 2} (No Atoms Mismatch)
\begin{flalign}\label{A3}
\nonumber
|\mathbf{a}^T_i \mathbf{d}_j| 
& \leq \frac{1}{\sigma_\mathrm{y}[i]} |\mathbf{d}^T_i \mathbf{d}_j| \\
& \leq \frac{\mu}{\sigma_\mathrm{y}[i]} \quad \forall \{ (i,j) : i\neq j, i \in K, j \notin K \}.
\end{flalign}
In other words, RFN can only decrease local stripe correlation to dictionary atoms that are not on the support. This assumption is very accurate, even in practical cases.

\textbf{A 3} (Noise Scaling)
\begin{equation}\label{A4}
|\mathbf{a}^T_i \mathbf{e}| \leq |\mathbf{W}_0\mathbf{d}^T_i \mathbf{e}| 
\leq \frac{\|\mathbf{d}^T_i\|_2 \|\mathbf{e}_i\|_2}{\min_i \sigma_\mathrm{y}[i]} \leq  \frac{\varepsilon_{\mathrm{d}}}{\tau} = \varepsilon_\mathrm{s} \quad \forall i,
\end{equation}
where $\mathbf{e}_i$ is a noise stripe of length $L_\mathrm{h}$ such that $\|\mathbf{e}_i\|_2=\varepsilon_{\mathrm{d}}$ ,
based on Cauchy-Schwartz and assuming atoms with unit norm $\mathbf{d}^T_i \mathbf{e} \leq \|\mathbf{d}^T_i\|_2 \|\mathbf{e}_i\|_2 = \|\mathbf{e}_i\|_2$.
In other words, the correlation of a local stripe noise to the dictionary atoms after RFN remains well bounded. Note that $\varepsilon_\mathrm{d} << \varepsilon$.

One may suggest that the above assumptions simply propose that thresholding the expression $\mathbf{D}^T\mathbf{W}_0\mathbf{y}$ is equivalent to thresholding $\mathbf{W}_0\mathbf{D}^T\mathbf{y}$. In other words, scaling the projection values is equivalent (or advantageous) to normalizing the (error) signal and then projecting it on the dictionary. In practice the two approaches are mathematically equivalent only when the atoms are at least $L_\mathrm{s}$ samples apart. Indeed, projection post-normalization can distort the observed signal. However, it is beneficial for preventing false detections. For example, in Section \ref{sec5} we presented examples using the Ricker wavelet. RFN helps to prevent the ``side wings" from being misdetected as main lobes of the wavelet. This is achieved by the neighborhood energy reducing the normalized side lobe's correlation to atoms at these locations, implying that it cannot be a true support location. 

These three assumptions are not necessarily realistic in the case of the seismic inversion. However many assumptions made for seismic deconvolution in the field of seismic processing are not necessarily realistic either, and yet seismic deconvolution has been applied for many years with considerable success \cite{Yilmaz:2001}.

Now we shall proceed to the proof of Theorem~\ref{Theorem 1}. Perfect support detection is guaranteed if the requirement
\begin{equation}\label{A5}
\min_{i \in K} |\mathbf{a}_i^T\mathbf{y}| > \max_{j \notin K} |\mathbf{a}_j^T\mathbf{y}|
\end{equation}
is met.
Assuming a rectangular receptive field normalization window of length $L_\mathrm{h}=L_\mathrm{d}$.
\begin{equation}\label{A6}
\sigma_\mathrm{y}[i] = \| \mathbf{D}_i \mathbf{x}_i + \mathbf{e}_i\| _2,
\end{equation}
where $\mathbf{x}_i$ denotes a stripe of length $m L_\mathrm{s}$ located around a data point $i$, and $\mathbf{D}_i$ is the submatrix partial dictionary yielding point $\mathbf{y}[i]$, that is obtained by restricting the dictionary $\mathbf{D}$ to the support of $m L_\mathrm{s}$ weights corresponding to $L_\mathrm{h}$ data points equally distributed around point $i$.
Denote $\mu = \mu(\mathbf{D}_i)$ (obviously $\mu = \mu(\mathbf{D})$), and assume without loss of generality the dictionary columns are normalized, that is 
$\sigma_{\mathrm{d}_i}=1 \forall i \in [1,M]$, 
according to the stripe Restricted Isometry Property (sRIP) (\cite[Definition 14]{Papyan:2017A})
\begin{equation}\label{A7}
 \Big(1-(s-1)\mu \Big) \ \| \mathbf{x}_i \|^2_2 \leq \| \mathbf{D}_i \mathbf{x}_i \|^2 _2 \leq \Big(1+(s-1)\mu \Big) \ \| \mathbf{x}_i \|^2_2.
\end{equation}
Moreover,
\begin{equation}\label{A8}
\| \mathbf{D}_i \mathbf{x}_i \|_2  \leq \sum_{j \in S^h_i} \| \mathbf{d}_j \|_2 \| \mathbf{x}_i \|_2 \leq \| \mathbf{x}_i \|_2 .
\end{equation}
Then, we can lower-bound the right term in (\ref{A5}) by
\begin{flalign}\label{A9}
\nonumber
\min_{i \in K} |\mathbf{a}_i^T\mathbf{y}| & 
\geq 
\min_{i \in K} |\mathbf{a}_i^T(\mathbf{y}-\mathbf{e})| - |\mathbf{a}_i^T\mathbf{e}| \\
\nonumber
&
\geq
\min_{i \in K} \frac{|\mathbf{d}_i^T \mathbf{D} \mathbf{x}|}{\|\mathbf{D}_i \mathbf{x}_i + \mathbf{e}_i\|_2} - \varepsilon_\mathrm{s} \\
\nonumber
&
\geq
\min_{i \in K} \frac{|\mathbf{d}_i^T\mathbf{D}_i \mathbf{x}_i|}{\|\mathbf{D}_i \mathbf{x}_i\|_2 + \|\mathbf{e}_i\|_2} - \varepsilon_\mathrm{s} 
\\
\nonumber
&
\geq
\min_{i \in K} \frac{|\mathbf{d}_i^T \mathbf{D}_i \mathbf{x}_i|}{\|\mathbf{x}_i\|_2 + \|\mathbf{e}_i\|_2} -  \varepsilon_\mathrm{s} 
\\ 
\nonumber
& \geq
\min_{i \in K} \frac{|\mathbf{d}_i^T \mathbf{D}_i \mathbf{x}_i|}{\|\mathbf{x}_i\|_1+\varepsilon_{\mathrm{d}}} - \varepsilon_\mathrm{s} \\
\nonumber
& \geq
\min_{i \in K} \frac{|\mathbf{x}[i]| \mathbf{d}_i^T\mathbf{d}_i }{\| \mathbf{x}_i\|_1 + \varepsilon_{\mathrm{d}}} -\Bigg| \sum_{t \in S^i_h, t \neq i} \frac{x[t] \mathbf{d}_i^T\mathbf{d}_t }{\|\mathbf{x}_i\|_1 + \varepsilon_{\mathrm{d}}} \Bigg| - \varepsilon_\mathrm{s} 
\\ 
\nonumber
& \geq
\min_{i \in K} \frac{(1+\mu)|\mathbf{x}[i]|}{\|\mathbf{x}_i\|_1 + \varepsilon_{\mathrm{d}}} - \mu \Big( \sum_{t \in S^i_h} \frac{|\mathbf{x}[t]|}{\|\mathbf{x}_i\|_1}  \Big) - \varepsilon_\mathrm{s}
\\
& \geq
\min_{i \in K} \frac{(1+\mu)|\mathbf{x}[i]|}{\|  \mathbf{x}_i\|_1 + \varepsilon_{\mathrm{d}}} - \mu - \varepsilon_\mathrm{s},
\end{flalign}
where we have used the relation $\|\mathbf{x}_i\|_1 \geq \|\mathbf{x}_i\|_2$ and assumptions \textbf{A 1}-\textbf{A 3}.
On the other hand,
\begin{flalign}\label{A10}
\nonumber
\max_{j \notin K} |\mathbf{a}_j^T\mathbf{y}| 
& = 
\max_{j \notin K} \frac{|\mathbf{d}_j^T\mathbf{y}|}{\|\mathbf{D}_j \mathbf{x}_j + \mathbf{e}_j \|_2} + |\mathbf{a}_j^T\mathbf{e}|\\
\nonumber
& \leq
\frac{\mu \|\mathbf{x}_j \|_1}{\|\mathbf{x}_j\|_2\sqrt{1-(s-1)\mu} - \varepsilon_{\mathrm{d}} } + \varepsilon_\mathrm{s}
\\
\nonumber
& \leq
\frac{\sqrt{s} \mu \|\mathbf{x}_j \|_2}{\|\mathbf{x}_j\|_2 \big(\sqrt{1 -(s-1)\mu} - \tilde{\varepsilon}_{\mathrm{d}} \big) } + \varepsilon_\mathrm{s} \\
& \leq
\frac{\sqrt{s} \mu}{\sqrt{1 -(s-1)\mu} - \tilde{\varepsilon}_{\mathrm{d},\infty}} + \varepsilon_\mathrm{s}.
\end{flalign}
where we denote $\tilde{\varepsilon}_{\mathrm{d}}= \frac{\varepsilon_{\mathrm{d}}}{\|\mathbf{x}_j\|_2}$, and 
$\tilde{\varepsilon}_{\mathrm{d},\infty} = \frac{\varepsilon_{\mathrm{d}}}{|\mathbf{x}|_\mathrm{min}}$.
Combining (\ref{A9}) and (\ref{A10}) we require
\begin{equation}\label{A11}
\min_{i \in K} \frac{|\mathbf{x}[i]|(1+\mu)}{\|\mathbf{x}_i\|_1 + \varepsilon_{\mathrm{d}}}  - \mu - \varepsilon_\mathrm{s} 
>
\frac{\sqrt{s} \mu}{\sqrt{1 -(s-1)\mu}-\tilde{\varepsilon}_{\mathrm{d},\infty}} + \varepsilon_\mathrm{s}.
\end{equation}
It follows that
\begin{flalign}\label{A12}
\nonumber
\frac{|\mathbf{x}[i]|}{\|\mathbf{x}_i\|_1 + \varepsilon_{\mathrm{d}}}  > \frac{\mu}{1+\mu} \Bigg( 1 +\frac{\sqrt{s}}{\sqrt{1 -(s-1)\mu}-\tilde{\varepsilon}_{\mathrm{d},\infty}} \Bigg) + \frac{2\varepsilon_\mathrm{s}}{1+\mu} \\
 \quad \forall i \in K.
\end{flalign}
Taking the worst case analysis point of view, considering that $\|\mathbf{x}_i\|_1 \leq s |\mathbf{x}_i|_\mathrm{max}$, we deduce
\begin{flalign}
\nonumber
\hspace{-5pt}
\frac{|\mathbf{x}_i|_\mathrm{min}}{|\mathbf{x}_i|_\mathrm{max}+\frac{\varepsilon_{\mathrm{d}}}{s}} > \frac{s\mu}{1+\mu} \Bigg( 1 +\frac{\sqrt{s}}{\sqrt{1 -(s-1)\mu}-\tilde{\varepsilon}_{\mathrm{d},\infty} } \Bigg) + \frac{2 s \varepsilon_\mathrm{s}}{1+\mu} 
\\
\forall i \in K.
\end{flalign}
In a noise-free model we have
\begin{equation}\label{A13}
\frac{|\mathbf{x}_i|_\mathrm{min}}{|\mathbf{x}_i|_\mathrm{max}}  > \frac{s\mu}{1+\mu} \Bigg( 1 +\frac{\sqrt{s}}{\sqrt{1 -(s-1)\mu} } \Bigg) \quad \forall i \in K.
\end{equation}
This completes the proof.
\qed

\section*{Appendix B}
\textbf{Proof of Theorem 2}

Perfect support detection is guaranteed if the requirement
\begin{equation}
\min_{i \in K} |\mathbf{a}_i^T\mathbf{y}| > \max_{j \notin K} |\mathbf{a}_j^T\mathbf{y}| \tag{\ref{A5}}
\end{equation}
is met.
When $s=1$ assuming $\mathbf{y}=\mathbf{Dx}$, we have
\begin{flalign}\label{B2}
\nonumber
\min_{i \in K} |\mathbf{a}_i^T\mathbf{y}| & = 
\min_{i \in K} \frac{|\mathbf{d}_i^T\mathbf{y}|}{\|\mathbf{D}_i \mathbf{x}_i\|_2} \geq
\min_{i \in K} \frac{|\mathbf{d}_i^T\mathbf{y}|}{\|\mathbf{x}_i\|_2} 
\\ 
& \geq
\min_{i \in K} \frac{|\mathbf{x}[i]|}{|\mathbf{x}[i]|} \mathbf{d}_i^T\mathbf{d}_i = 1. 
\end{flalign}
on the other hand,
\begin{flalign}\label{B3}
\nonumber
\max_{j \notin K} |\mathbf{a}_j^T\mathbf{y}| & =
\max_{j \notin K} \frac{|\mathbf{d}_j^T\mathbf{y}|}{\|\mathbf{D}_j \mathbf{x}_j\|_2} \\
&
\leq
\frac{\mu |\mathbf{x}_j[k_j]| }{|\mathbf{x}_j[k_j]| }
\leq
\mu, \quad k_j \in K, k_j \in S^j_h.
\end{flalign}
Since $\mu<1$, for any $0 <\frac{|\mathbf{x}|_{\mathrm{min}}}{|\mathbf{x}|_{\mathrm{max}}} \leq 1$ the support is perfectly recovered at the first iteration using RFN.
\qed

\section*{Appendix C}
\textbf{Proof of Theorem 3}

Recall that we have defined the stripe local energy in (\ref{3.25}) as
\begin{equation}\label{C1}
\sigma_{\mathrm{y}}[k] = \| \mathbf{H} \mathbf{y}_k \|_2 = \| \mathbf{HD}_k \mathbf{x}_k \|_2.
\end{equation} 
Similarly to the stripe Restricted Isometry Property (sRIP) (\cite[Definition 14]{Papyan:2017A})
we can observe that
\begin{flalign}\label{C2}
\nonumber
\| \mathbf{HD}_i \mathbf{x}_i \|^2 _2 & =
\mathbf{x}_i^T (\mathbf{HD}_i)^T \mathbf{HD}_i \mathbf{x}_i 
\\
\nonumber
&
\leq 
|\mathbf{x}_i|^T [\mathbf{H}^{2}_\mathrm{d}+\mu(\mathbf{HD}_i)(\mathbf{1}-\mathbf{I})] |\mathbf{x}_i|
\\
\nonumber
&
\leq 
\ \| \mathbf{H}_\mathrm{d} \mathbf{x}_i \|^2_2 - \mu(\mathbf{HD}_i) \|\mathbf{x}_i \|^2_2 + \mu(\mathbf{HD}_i) \|\mathbf{x}_i \|^2_1
\\
&
\leq 
\ \| \mathbf{H}_\mathrm{d} \mathbf{x}_i \|^2_2 + (s-1)\mu(\mathbf{HD}_i)\|\mathbf{x}_i \|^2_2,
\end{flalign}
where $\mathbf{H}_\mathrm{d}$ is a diagonal matrix holding the attenuated atoms $\ell_2$ norms, namely $\mathbf{H}^2_\mathrm{d} = \mathrm{diag}(\mathrm{trace}((\mathbf{HD}_i)^T \mathbf{HD}_i))$ and $\mathbf{H}_\mathrm{d}[j,j]=\|\mathbf{Hd}^i_j\|_2, \ j \in S^h_i$, where $\mathbf{d}^i_j \in \mathbb{R}^{L_{\mathrm{h}} \times 1}$ is the $j$th atom of the submatrix $\mathbf{D}_i$.
Clearly, we also have
\begin{flalign}\label{C3}
\nonumber
\| \mathbf{HD}_i \mathbf{x}_i \|^2 _2 & =
\mathbf{x}_i^T (\mathbf{HD}_i)^T \mathbf{HD}_i \mathbf{x}_i 
\\
\nonumber
&
\geq 
|\mathbf{x}_i|^T [\mathbf{H}^{2}_\mathrm{d}-\mu(\mathbf{HD}_i)(\mathbf{1}-\mathbf{I})] |\mathbf{x}_i|
\\
\nonumber
&
\geq 
\ \| \mathbf{H}_\mathrm{d} \mathbf{x}_i \|^2_2 + \mu(\mathbf{HD}_i) \|\mathbf{x}_i \|^2_2 - \mu(\mathbf{HD}_i) \|\mathbf{x}_i \|^2_1
\\
&
\geq 
\ \| \mathbf{H}_\mathrm{d} \mathbf{x}_i \|^2_2 - (s-1)\mu(\mathbf{HD}_i)\|\mathbf{x}_i \|^2_2.
\end{flalign}
By assumption $h$ is a receptive field normalization kernel that is monotonically descending away from its center, therefore we assume $\mu(\mathbf{HD}_i) << \mu(\mathbf{D}_i)$, and hereafter use the approximation
\begin{equation}\label{C4}
\sigma_{\mathrm{y}}[k] \approx \| \mathbf{H}_\mathrm{d} \mathbf{x}_i \|_2.
\end{equation} 
Also, under the separation condition let us denote $h_\mathrm{d}(\nu) \triangleq \max_{p \in [1,m]} \mathbf{H}_\mathrm{d}[\Delta_k+(p-1)L_\mathrm{s},\Delta_k+(p-1)L_\mathrm{s}]$. In other words $h_\mathrm{d}(\nu)$ is the maximal $\ell_2$ norm of an atom shifted by the separation distance multiplied by the RFN window.

As stated above, to guarantee perfect support detection we have to show that the requirement 
\begin{equation}
\min_{i \in K} |\mathbf{a}_i^T\mathbf{y}| > \max_{j \notin K} |\mathbf{a}_j^T\mathbf{y}| \tag{\ref{A5}}
\end{equation}
is satisfied.
We shall begin with the left-hand-side term,
\begin{flalign}\label{C5}
\nonumber
\hspace{-3pt}
\min_{i \in K} |\mathbf{a}_i^T\mathbf{y}| & 
\geq
\min_{i \in K} \frac{|\mathbf{d}_i^T \mathbf{D} \mathbf{x}|}{\|\mathbf{HD}_i \mathbf{x}_i\|_2} 
\approx
\min_{i \in K} \frac{|\mathbf{d}_i^T \mathbf{D} \mathbf{x}|}{\|\mathbf{H}_\mathrm{d}\mathbf{x}_i\|_2} \\
\nonumber
& \geq
\min_{i \in K} \frac{|\mathbf{x}[i]|}{\|\mathbf{H}_\mathrm{d}\mathbf{x}_i\|_2} \mathbf{d}_i^T\mathbf{d}_i -\Bigg| \sum_{t \in S^i_h, t \neq i} \frac{x[t]}{\|\mathbf{H}_\mathrm{d}\mathbf{x}_i\|_2} \mathbf{d}_i^T\mathbf{d}_t \Bigg|
\\ 
\nonumber
& \geq
\min_{i \in K} \frac{|\mathbf{x}[i]|- \mu \sum_{t \in S^i_h, t \neq i} |\mathbf{x}[t]|}{\|\mathbf{H}_\mathrm{d}\mathbf{x}_i\|_1}
\\
\nonumber
& \geq
\min_{i \in K} \frac{|\mathbf{x}[i]|- \mu \sum_{t \in S^i_h, t \neq i} |\mathbf{x}[t]|}
{|\mathbf{x}[i]|+ h_\mathrm{d}(\nu) \sum_{t \in S^i_h, t \neq i} |\mathbf{x}[t]|}
\\
& 
=
1 - \big( \mu + h_\mathrm{d}(\nu) \big) \max_{i \in K} \frac{\sum_{t \in S^i_h, t \neq i} |\mathbf{x}[t]|}
{|\mathbf{x}[i]|+ h_\mathrm{d}(\nu) \sum_{t \in S^i_h, t \neq i} |\mathbf{x}[t]|},
\end{flalign}
where we have used $ \|\mathbf{H}_\mathrm{d}\mathbf{x}_i\|_1 \leq |\mathbf{x}[i]|+ h(\nu) \sum_{t \in S^i_h, t \neq i} |\mathbf{x}[t]|$ since the closest nonzero input entry is at distant $\Delta_k$ samples from $i$.
Turning now to the right-hand-side of (\ref{A5}) we have
\begin{flalign}\label{C6}
\nonumber
\max_{j \notin K} |\mathbf{a}_j^T\mathbf{y}| 
& = 
\max_{j \notin K} \frac{|\mathbf{d}_j^T\mathbf{y}|}{\|\mathbf{HD}_j \mathbf{x}_j\|_2} 
\\
&
\leq
\frac{\mu \|\mathbf{x}_j \|_1}{\|\mathbf{H}_\mathrm{d}\mathbf{x}_j\|_2 }
\leq
\frac{\mu \|\mathbf{x}_j \|_1}{\frac{h_\mathrm{d,min}}{\sqrt{s}}\|\mathbf{x}_j\|_1}
\leq
\frac{\sqrt{s} \mu}{h_\mathrm{d,min}},
\end{flalign}
where we have lower bounded the denominator, assuming $\sigma_{\mathrm{y}}[j]>\tau$,
\begin{equation}\label{C7}
\|\mathbf{H}_\mathrm{d}\mathbf{x}_j\|_2
\geq
{h_\mathrm{d,min}}\|\mathbf{x}_j\|_2
\geq
\frac{{h_\mathrm{d,min}}}{\sqrt{s}} \|\mathbf{x}_j\|_1,
\end{equation}

It is worth mentioning that a tighter bound could be obtained using
\begin{flalign}\label{C6b}
\nonumber
\max_{j \notin K} |\mathbf{a}_j^T\mathbf{y}| 
& = 
\max_{j \notin K} \frac{|\mathbf{d}_j^T\mathbf{y}|}{\|\mathbf{HD}_j \mathbf{x}_j\|_2}
\\
& 
\leq
\frac{\sqrt{s} \sum_{k \in S_j^h,k \neq j} |\mathbf{d}_j^T\mathbf{d}_k||\mathbf{x}_j[k] |}
{\sum_{k \in S_j^h,k \neq j} \mathbf{H}_\mathrm{d}[|k-j|] |\mathbf{x}_j[k]|}.
\end{flalign}
Which means no false detection are guaranteed as long as
\begin{equation}\label{C8b}
\frac{ \sqrt{s} |\mathbf{d}_j^T\mathbf{d}_k| }{\mathbf{H}_\mathrm{d}[|k-j|]} < \beta_1, \forall k \in S_j^h, j \notin K.
\end{equation}

Combining (\ref{C7}) and (\ref{C5}), and denote $\| \mathbf{x}_{-i} \|_1 \triangleq \sum_{t \in S^i_h, t \neq i} |\mathbf{x}[t]| = \|\mathbf{x}_i\|_1 - |\mathbf{x}[i]| $, we have that
as long as
\begin{equation}\label{C8}
\frac{\sqrt{s} \mu}{h_\mathrm{d,min}} < \beta_1,
\end{equation}
and 
\begin{equation}\label{C9}
\frac{\| \mathbf{x}_{-i} \|_1}{|\mathbf{x}[i]| + h_\mathrm{d}(\nu)\| \mathbf{x}_{-i} \|_1} < \frac{1-\beta_1}{\mu+h_\mathrm{d}(\nu)} \quad \forall i \in K,
\end{equation}
the true support is fully recovered by RFN thresholding.
This concludes the proof. \qed

\section*{Appendix D}
\textbf{Earth Q-model time-variant wavelets estimation}

We recall from \cite{Pereg:2017A,Pereg:2019A}, that we can estimate the set of time-variant pulses $\{g_{\sigma,n}\}$ in (2.2) according to the earth Q-model.
Namely, we begin with an initial wavelet with a dominant frequency $\omega_0$, e.g., the real-valued Ricker wavelet,
\begin{equation}
	g(t) = \Big(1-\frac{1}{2}\omega_0^2t^2\Big)\exp\Big(-\frac{1}{4}\omega_0^2t^2\Big).
	\tag{\ref{5.1}}
\end{equation}
In this setting, the scaling parameter is $\sigma=\omega_0^{-1}$. Following the earth Q-model \cite{Wang1}, assuming we know $Q$ \cite{Kjartansson}, a reflected pulse at travel time $t_n$, is 
\begin{equation}\label{D.2}
	u_n(t-t_n) = \text{Re} \Big\{ \frac{1}{\pi}\int_{0}^{\infty} G(\omega)\exp[j(\omega t-\kappa r(\omega))]d\omega \Big\},
\end{equation}
where $G(\omega)$ is the Fourier transform of the source waveform $g(t)$,
\begin{equation}\label{D.3}
	\kappa r(\omega)\triangleq \Big(1-\frac{j}{2Q}\Big)\left|\frac{\omega}{\omega_0}\right|^{-\gamma} \omega t_n,
\end{equation}
\begin{equation}
	\gamma \triangleq \frac{2}{\pi}\tan^{-1}\Big(\frac{1}{2Q}\Big)\approx\frac{1}{\pi Q},
\end{equation} 
Note that in the frequency domain, the phase change exponential operator represents velocity dispersion, while the amplitude attenuation exponential operator corresponds to energy absorption of the traveling pulses
\begin{flalign}\label{D.4}
\nonumber
	U_{n}(\omega)\exp^{-j \omega t_n} & = G(\omega)\exp\Big(\	-j \left| \frac{\omega}{\omega_0}\right|^{-\gamma} \omega t_n \Big)\\
& \times \exp\Big(-\left|\frac{\omega}{\omega_0}\right|^{-\gamma} \frac{\omega t_n}{2Q} \Big).
\end{flalign}
The time-domain seismic pulse reflected at two-way travel time (depth) $t_n$ is
\begin{equation}\label{D.5}
			u_n(t-t_n)=\frac{1}{2\pi}\int U_{n}(\omega)\exp[j \omega (t - t_n)] d\omega.
\end{equation}
Therefore, the estimated set of pulses $\{g_{\sigma,n}\}$ based on the earth Q model are defined as
\begin{equation}\label{2.8}
g_{\sigma,n}(t-t_n)= u(t-t_n)|_{\sigma=\omega_0^{-1}}.
\end{equation}

\bibliography{seismic_cnn}



\end{document}